\pdfoutput=1
\documentclass[11pt]{article}

\usepackage[preprint]{acl}

\usepackage{times}
\usepackage{latexsym}
\usepackage{amsmath}
\usepackage{dsfont}
\usepackage{adjustbox}

\usepackage[T1]{fontenc}
\usepackage[utf8]{inputenc}

\usepackage{microtype}

\usepackage{inconsolata}

\usepackage{graphicx}

\usepackage{booktabs}
\usepackage{multirow}
\usepackage{multicol}
\usepackage{makecell}
\usepackage{subcaption}
\usepackage{todonotes}
\usepackage{amsfonts}
\usepackage{enumitem}
\usepackage{float}

\newcommand{\narrowbotc}[1]{{\colorbox{yellow}{\parbox[t][][t]{23em}{#1}}}}
\newcommand{\na}{\ensuremath{\text{N/A}}}
\newcommand{\skipped}{\ensuremath{\text{N/S}}}

\title{Large Language Models Do Multi-Label Classification Differently}
\author{
 \textbf{Marcus Ma}\thanks{Equal contribution. Code is available at \url{https://github.com/gchochla/LLM-multilabel-differently}.},
 \textbf{Georgios Chochlakis}\footnotemark[1],
 \textbf{Niyantha Maruthu Pandiyan},\\
 \textbf{Jesse Thomason},
 \textbf{Shrikanth Narayanan}
\\
 University of Southern California
 \\
 \small{
   \textbf{Correspondence:} \{\href{mailto:mjma@usc.edu}{mjma}, \href{mailto:chochlak@usc.edu}{chochlak}\}@usc.edu
 }
}

\begin{document}
\maketitle
\begin{abstract}

% Multi-label language tasks have been largely overlooked in the age of LLMs, even though they represent more realistic classification settings. In this work, we investigate how LLMs perform multi-label classification. We find that LLMs need multiple steps to generate all the predictions, as they produce unrepresentative distributions at each generation step, caused by language modeling interfering with the classification task. Specifically, LLMs generate spiky probability distributions, yielding one label at a time with high confidence, and suppressing the probabilities of the other labels. Once a label is generated, a new probability distribution spikes for the next predicted label. We also observe that, as scale increases, the output distributions become spikier, but the relative ordering of labels within these distributions improves. Moreover, as in single-label settings, we find that finetuning (e.g., SFT, RLHF) further increases the spikiness of model outputs. We then introduce the task of distribution alignment to the multi-label setting between LLM-derived label probability distributions and a human distribution, approximated from annotator samples in subjective tasks. We propose novel methods for both zero-shot and supervised settings that improve upon both alignment and performance over existing approaches.

Multi-label classification is prevalent in real-world settings, but the behavior of Large Language Models (LLMs) in this setting is understudied. We investigate how autoregressive LLMs perform multi-label classification, focusing on subjective tasks, by analyzing the output distributions of the models at each label generation step. We find that the initial probability distribution for the first label often does not reflect the eventual final output, even in terms of relative order and find LLMs tend to suppress all but one label at each generation step. We further observe that as model scale increases, their token distributions exhibit lower entropy and higher single-label confidence, but the internal relative ranking of the labels improves. Finetuning methods such as supervised finetuning and reinforcement learning amplify this phenomenon. We introduce the task of distribution alignment for multi-label settings: aligning LLM-derived label distributions with empirical distributions estimated from annotator responses in subjective tasks. We propose both zero-shot and supervised methods which improve both alignment and predictive performance over existing approaches. We find one method -- taking the max probability over all label generation distributions instead of just using the initial probability distribution -- improves both distribution alignment and overall F1 classification without adding any additional computation.
\end{abstract}

\section{Introduction}
\label{sec:intro}

% The majority of language tasks operate under the assumption that each document maps to a single, unambiguous ground truth. However, in real-world scenarios, the truth is rarely obvious and singular, unless categories are mutually exclusive. Multi-label formulations, where none, one, or multiple labels can be assigned to each document, enable us to better capture the complexity for each task more flexibly, not just through the presence of multiple categories, but also by the inter-label correlations. Moreover, degrees of belief for each label may play an important modeling role. However, little research has been conducted on multi-label modeling with Large Language Models (LLMs).

Many natural language processing tasks assume each input has a single, unambiguous label, represented as a one-hot encoding~(\citealt{srivastava2022beyond, wang_2024}; inter alia). However, in realistic settings, especially where categories are not mutually exclusive, this assumption fails. Multi-label classification, where instances can have none, one, or multiple labels, better captures the inherent ambiguity, richness of human categorization, and label correlations, notably in subjective tasks~\cite{mohammad2018semeval, demszky2020goemotions}. It also enables modeling degrees of belief, which is integral in subjective tasks to express confidence or intensity in each label~\cite{paletz2023social}. Intensity is a tool not generally available in single-label settings. Despite their widespread applicability, multi-label tasks have received little attention in the context of Large Language Models (LLMs).

\begin{figure}[!t]
    \centering
    \includegraphics[width=1\linewidth]{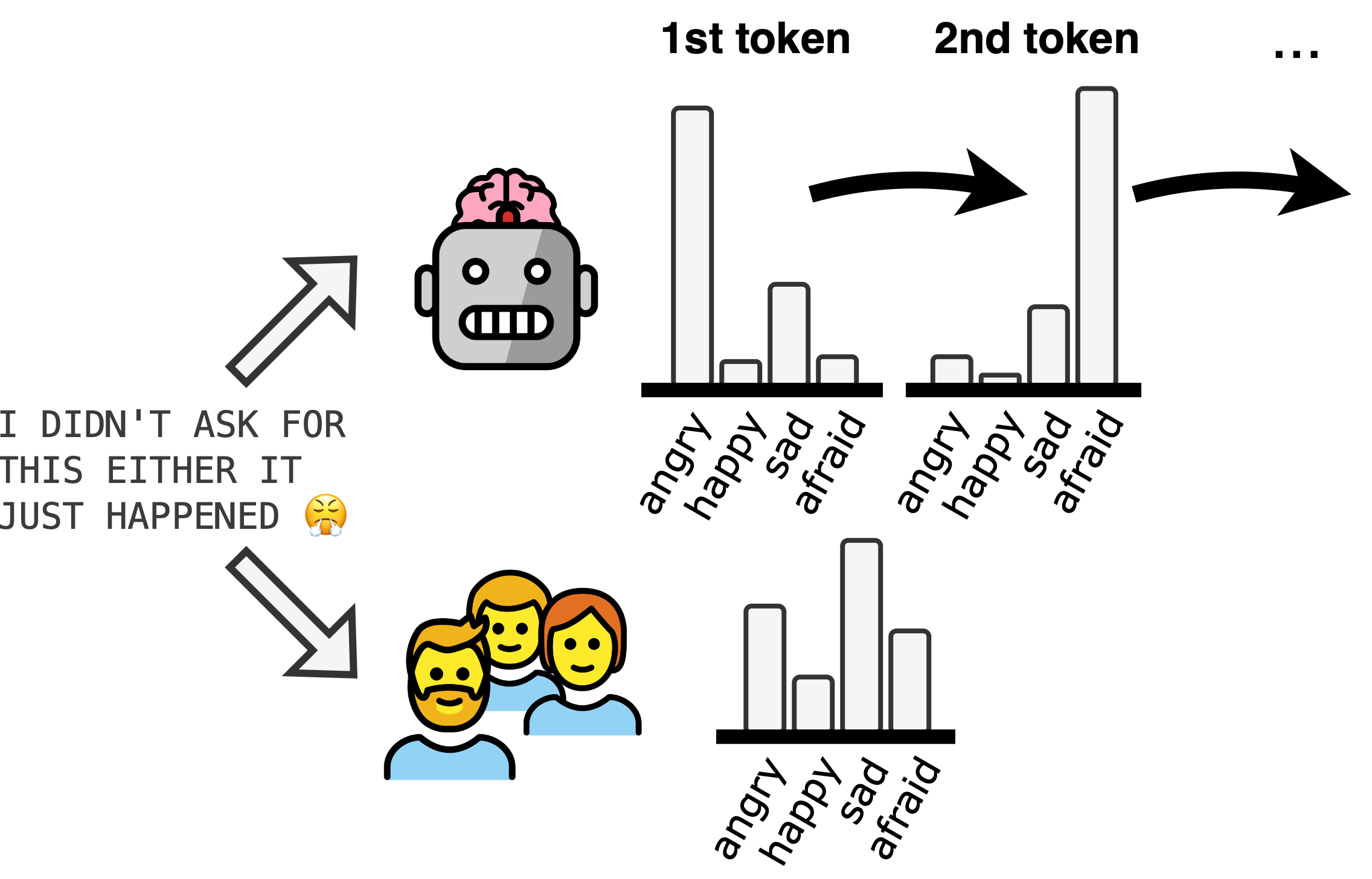}
    \caption{Autoregressive language modeling is incompatible and interferes with multi-label classification: LLMs generate one label at a time with unrepresentative distributions misaligned from reference distributions.}
    \label{fig:thumbnail}
\end{figure}

% \begin{figure}[!t]
%     \centering
%     \includegraphics[width=1\columnwidth]{figs/spikiness.png}
%     \caption{Sorted label probabilities when generating the first label for Llama3 70B Instruct. Most distributions are spiky, with the top label having near-1 probability.}
%     \label{fig:spikiness}
% \end{figure}

% One potential explanation is that single-label tasks better fit the language modeling objective: LLMs are trained to produce discrete probability distributions via joint normalization (softmax) of logits over vocabulary tokens. These can be readily used in a single-label setting by normalizing only across the label tokens of interest for the task.

% In contrast, there is no direct way to apply this to the multi-label case. To allow for the presence of multiple labels, the label probabilities in this setting do not need to sum to $1$ and, consequently, do not form a proper discrete distribution. Notably, LLM logits have been trained to be normalized together, so their values have meaning in the context of the rest of the logits, and not in isolation. Normalizing the logits independently to derive probabilities then does not have any behavioral guarantees.

A key reason may be the incompatibility between the language modeling objective and the multi-label setting. LLMs are trained to generate probability distributions over vocabulary tokens via softmax normalization for the immediate next token, naturally lending themselves to single-label settings, such as by restricting the normalization to label tokens. In contrast, multi-label classification does not require label probabilities to sum to one. Instead, each label’s confidence can, in principle, be modeled independently. This runs counter to how LLMs are trained, as their logits are meaningful only in relation to each other.

% One possibility is that the confidence of LLMs in each label is still reflected in their relative probabilities. Although this still does not allow us to directly extract a proper multi-label distribution, we would be able derive multi-label predictions via thresholding~\cite{he2018joint}. This is not desirable in settings where degrees of belief can be expressed in the gold labels, and so the value for each is not confined to $\{0, 1\}$ but rather $[0, 1]$.

% Another straightforward approach is to allow the LLM to generate the entire sequence of predictions, incurring additional inference costs. But it is unclear how representative the sequence of discrete distributions is of the model's confidence for each label, as the logits still need to be normalized in the presence of the other labels. For instance, if the model's ``true'' confidence for one label is $0.6$, and otherwise zero for the rest of the labels, then the model would still need to assign non-zero probabilities for the rest of the labels to provide a score of $0.6$. Moreover, the consistency of these distributions across steps is questionable, as the model at each step is conditioned on its previous predictions.

Relative probabilities might still encode relevant information, useful for threshold-based prediction~\cite{he2018joint}, but such methods are ill-suited for tasks involving graded or subjective judgments, where ground truth can lie in $[0,1]$, not just $\{0,1\}$. Alternatively, LLMs can be allowed to autoregressively generate a sequence of labels. However, the resulting distributions at each step are conditioned on earlier outputs and remain constrained by the same joint normalization, making them difficult to interpret as genuine model confidence scores~\cite{breen_2018}. For example, a model with 60\% confidence in a label still needs to allocate the remaining 40\% among competing options, regardless of its ``true'' confidence.

% In this work, we present our findings into the underlying mechanisms that guide multi-label classification in LLMs. We find that LLMs create spiky probability distributions--that is, they predict a single label with very high confidence while the rest of the label confidences are close to zero. Then, the same pattern is repeated for potential subsequent labels until the output is fully generated.
% Moreover, we find that the distributions across steps are not consistent; for example, the second largest probability is not predicted at all the majority of the time, even if the model continues generating predictions.
% One interpretation of this behavior is that LLMs perform sequential single-label classification, rather than creating a holistic probability distribution over all possible labels.
% In essence, we demonstrate that language modeling interferes with multi-label classification, raising questions about the appropriateness of LLMs in these settings.

In this work, we investigate how LLMs generate multi-label predictions by analyzing their output distributions in each generation step. We show that LLMs exhibit spiky distributions, where each consecutive step strongly favors a single label while suppressing others. This pattern produces a list of high-confidence individual predictions rather than a comprehensive probability distribution. Notably, these distributions lack consistency across steps: labels with high probability in earlier steps are rarely revisited in subsequent ones, even when the model continues generating labels, which suggests that LLMs are performing sequential single-label classification and not holistic multi-label reasoning.

% To evaluate the alignment of these distributions with the gold labels, we propose baselines, and zero-shot and supervised techniques to derive a single multi-label distribution from LLMs, where each label's probability reflects the model's confidence for that class~\cite{griffiths_2010, oaksford_1994}.
% To evaluate confidence, not just predictions, we focus mostly on subjective tasks, where we can derive degrees of belief in each label. To do so, we refrain from using traditional aggregation methods, which have been criticized recently as inappropriate for subjective settings~\cite{dutta2023modeling, chochlakis2024aggregation}. Rather, we leverage the inherent plurality of valid interpretations in subjective tasks~\cite{kahneman_1972, tenenbaum_2006, griffiths_2010, oaksford_1994}
% and use individual annotations to represent samples drawn from an underlying \textit{human distribution}, which we approximate with the percentage of annotators who assigned the label to each document.
% Under this formulation of the human distribution, we extend the task of \textbf{distribution alignment} --originally introduced as distributional inference in \citet{zhou_2022}-- to the multi-label setting.

To evaluate this phenomenon, we frame distributional alignment as a core task: aligning LLM-derived distributions with ground-truth distributions. To evaluate confidence, not just predictions, we also compare with empirical distributions derived from human annotator responses. Rather than relying on hard label agreement (e.g., majority vote), we embrace the plurality of human interpretations~\cite{kahneman_1972, tenenbaum_2006, griffiths_2010, aroyo2015truth} and approximate the distribution for each document by the empirical proportion of annotators selecting each label, resulting in values $\in[0, 1]$. We extend the distributional inference framework~\cite{zhou_2022} to the multi-label setting and evaluate both zero-shot and supervised approaches for aligning LLM outputs with the human-annotation derived empirical distributions.

% Our contributions are summarized as follows:
% \ma{could cut "contributions" list, if we need space; but I generally like having it in my papers}
% \begin{enumerate}
%     \item In \S\ref{sec:how_multi_label}, we formally investigate how LLMs generate multi-label predictions. We find that LLMs generate one label at a time, similar to the language modeling objective.
%     \item In \S\ref{sec:distribution_alignment}, we evaluate LLMs on the task of multi-label distributional alignment with an underlying, diverse ``human'' distribution.
%     \item In \S\ref{sec:results}, we present results from our experiments that show that our test-time and supervised methods outperform baselines.
% \end{enumerate}

Our contributions are the following:
\begin{itemize}[nosep,noitemsep]
\item In \S\ref{sec:how_multi_label}, we provide the first formal analysis of how LLMs handle multi-label classification, showing that their prediction behavior mirrors the steps inherent in the language modeling that favor a single-label setting.
\item In \S\ref{sec:distribution_alignment}, we introduce and evaluate distribution alignment in the multi-label setting, using degrees of belief as a reference distribution. We show that our proposed zero-shot and supervised methods improve alignment and predictive quality over standard baselines on subjective multi-label tasks.
\item We highlight the zero-shot approach of max-over-generations, which improves both distribution alignment and F1 classification for no additional computation. This method involves setting a label's probability to its max value across all label generations rather than its value in a single label distribution.
\end{itemize}

\section{Related Work}
\label{sec:related_work}

\subsection{LLM Usage for Multi-label Predictions}
Single-label problems have dominated both early (e.g., ImageNet; ~\citealt{deng2009imagenet}) and recent (BigBench; ~\citealt{srivastava2022beyond}) deep learning progress, despite the obvious limitations of single-label settings when the labels are not mutually exclusive. ImageNet~\cite{deng2009imagenet} as a benchmark, for instance, used the top-$k$ accuracy to evaluate models in order to deal with the potential simultaneous existence of multiple categories within each image, which was not reflected in the annotations.
Similarly, previous multi-label modeling attempts treated the task as single-label by using the general cross-entropy loss with a threshold to turn the prediction into a proper multi-label output~\cite{he2018joint}. Subsequent works switched to the binary cross-entropy loss, and tried to leverage the relationship between labels for additional supervision~\cite{he2018joint, alhuzaliSpanemoCastingMultilabel2021, chochlakisLeveragingLabelCorrelations2023}.

To the best of our knowledge, \citet{niraula_2024} is the only work to explicitly investigate LLM multi-label classification~\cite{chen_2022} in niche domains. \citet{betianu_2024} explored a multi-label framework for finetuning BERT and \citet{jung_2023} trained a classifier on top of T5 encodings directly for multi-label classification rather than relying on model text generation. The two well-studied forms of multi-label classification are extreme multi-label classification (XMLC; ~\citealt{zhu_2024}), where models must assign many labels to a document from a very large label set (1000+ labels), and hierarchical multi-label classification~\cite{tabatabaei_2025}, where labels are subdivided into sub-labels recursively. Subjective multi-label classification is relatively unexplored~\cite{chochlakis2024strong}. We thoroughly investigate LLMs in these settings by analyzing their classification patterns across datasets.

\subsection{Subjective Language Tasks}
Many works have attempted to model individual annotator perspectives and intensities~\cite{paletz2023social} instead of the majority vote, e.g., with EM~\cite{dawidMaximumLikelihoodEstimation1979a, hovy2013learning}, word embeddings~\citet{gartenIncorporatingDemographicEmbeddings2019}, and encoder-based approaches~\cite{gordonJuryLearningIntegrating2022, mokhberian2022noise, davaniDealingDisagreementsLooking2022, mokhberian2023capturing}. Modeling annotators with LLMs has shown limited success, and LLM biases have also been explored~\cite{dutta2023modeling, abdurahman2024perils, chochlakis2024aggregation}.

\begin{figure*}[!t]
     \centering
    \includegraphics[width=1\linewidth]{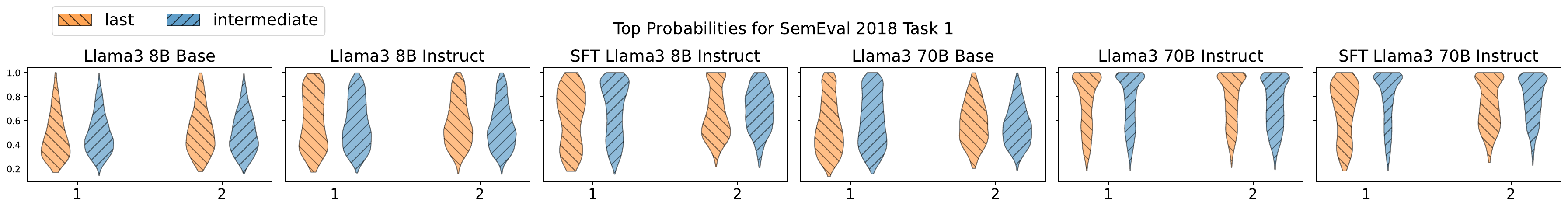}     
    \includegraphics[width=1\linewidth]{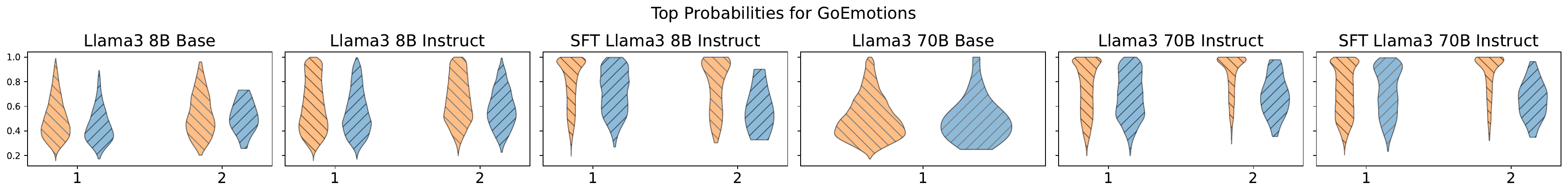}     
    \includegraphics[width=1\linewidth]{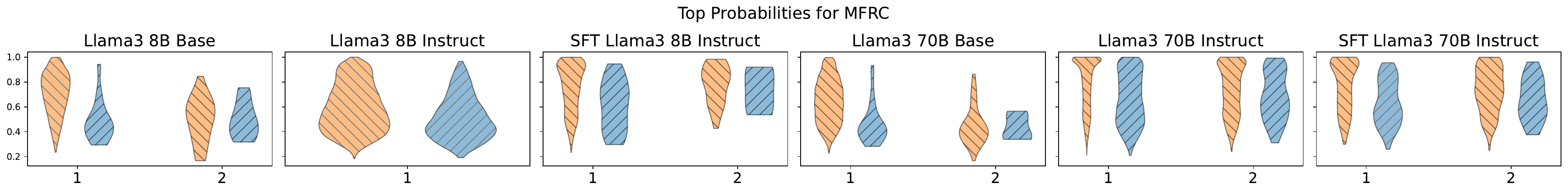}     
    \includegraphics[width=0.67\linewidth]{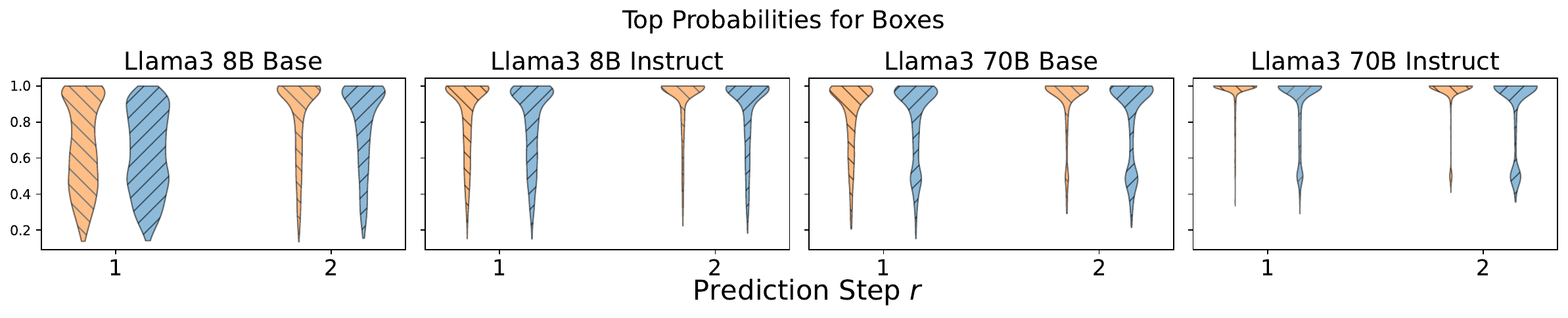}     

    \caption{Top probabilities at each generation step when the \texttt{last} or an \texttt{intermediate} label is generated. Patterns are identical between the two settings, and bigger or finetuned models have clusters closer to 100\%. A single step only is shown when only up to labels were generated for all examples in a specific setting.}
    \label{fig:top-violins}
\end{figure*}

\subsection{Calibration for LLMs}
Increasing the size of neural networks generally improves performance and generalization \cite{hoffmann_2022, brutzkus_2019, kaplan_2020}. However, while smaller models essentially produce well-calibrated predictions ``for free''~\cite{niculescu-mizil_2005}, as neural networks become increasingly complex, they are also less calibrated \cite{guo_2017}. Recent language models trained with Reinforcement Learning from Human Feedback (RLHF) have seen ``spiky'' probability distributions where models are overconfident in a select few output tokens while suppressing the probabilities of other options~\cite{xie_2024, leng_2025}. Instruction tuning also appears to reduce calibration over base models \cite{zhu_2023}. Several methods have been proposed to improve LLM calibration, including temperature scaling \cite{xie_2024, huang_2024}, adding calibration metrics as a learnable feature \cite{chen_2023}, and in-context prompting \cite{zhao_2024}. Our proposed distribution alignment setting differs from calibration in that it compares the probabilities over the entire label set whereas calibration only compares the predicted label probability to the ground truth.
\section{Datasets}
\label{sec:datasets}

We present both objective and subjective multi-label datasets. We use 10-shot prompts with Llama3~\cite{dubey2024llama} (more details in \S\ref{sec:appendix-implementation_details}). We apply softmax over initial label tokens to derive label probabilities at each step. It is well known that several different tokens can correspond to the same concept \cite{holtzman_2022}, such as ``happy'', ``Happy'', and `` happy'', and found that selecting the highest logit score across all same-concept tokens as a given label's logit value was the most effective way to capture model belief.

\paragraph{Boxes \cite{kim2023entity}} Entity tracking based on natural language description of ``box'' contents and ``move'' operations. Each box can contain none, one, or multiple objects. The dataset contains thousands of synthetic examples.

\paragraph{SemEval 2018 Task 1 E-c \cite{mohammad2018semeval}} Multi-label emotion recognition of 11 emotions. We use the English tweets. We refer to this as \textbf{SemEval}. Although it does not contain annotator labels, it has a frequent presence of multiple labels, allowing us to study the generation dynamics.

\paragraph{MRFC~\cite{trager2022moral}} Multi-label moral foundation corpus of six moral foundations.
3 annotators were assigned to each sample.

\begin{table}[t!]
\centering
\begin{adjustbox}{max width=\columnwidth}
\begin{tabular}{lccccccc}
\toprule
Dataset & \shortstack{Annotators\\ (per example)} & \shortstack{Cohen's\\Kappa} & \shortstack{0\\labels} & \shortstack{1\\label} & \shortstack{2\\labels} & \shortstack{3+\\labels} \\
\midrule
GoEmotions  & 81 (3.58) & 0.27 & 29\% & 62\% & 8\% & 1\% \\
MFRC        & 6 (2.99) & 0.21 & 78\% & 18\% & 3\% & $<$1\% \\
SemEval      & --  (--)       & --    & 1\%  & 13\% & 40\% & 46\% \\
\bottomrule
\end{tabular}
\end{adjustbox}
\caption{Annotation statistics and label distributions. The public release of SemEval does not include individual annotator labels, only aggregates.}
\label{tab:annotation_stats}
\end{table}

\paragraph{GoEmotions~\cite{demszky2020goemotions}} Multi-label emotion recognition benchmark of 27 emotions. For efficiency, we pool the emotions to seven emotions via hierarchical clustering (see \S\ref{sec:appendix-implementation_details}).
On average, 3.6 annotators were assigned to each sample.

% For each dataset, we create two testing sets: a "multi-label only" set, containing data that exclusively has multiple ground truth labels, which we use in \S\ref{sec:how_multi_label}; and a main testing set, which contains a uniform number of data across three label types (no label, single label, and multi-label) and annotator disagreements (no disagreement and has disagreement) for our experiments in \S\ref{sec:distribution_alignment}. For each test set we select 200 data points per dataset.
% \ma{we can move descriptions to appendix if needed?}

\section{Multi-Label Mechanisms of LLMs}

% \begin{figure*}[!h]
%      \centering
%     \includegraphics[width=1\linewidth]{figs/SemEval-all/top_probabilities.pdf}     
%     \includegraphics[width=1\linewidth]{figs/GoEmotions-all/top_probabilities.pdf}     
%     \includegraphics[width=1\linewidth]{figs/MFRC-all/top_probabilities.pdf}     
%     \includegraphics[width=0.67\linewidth]{figs/Boxes-all/top_probabilities.pdf}     

%     \caption{Top two probabilities at each generation step $r$ (up to two for brevity) when the \texttt{last} label is generated, or when an \texttt{intermediate} label is, shown for four datasets. For each dataset, the bottom row shows the top probability, and the top row the second highest probability, in addition to the probability at the current step of the label that was actually predicted in the next step ($r$+1 \texttt{pred}), and the probability at the next generation step of the second highest probability of the current step (\texttt{intermediate @} $r$+1). Also shown is the percentage of cases the second-highest probability label at $r$ and the prediction at $r$+1 were the \texttt{same}. Instances with only one shown generation step predicted only up to two labels.}
%     \label{fig:main-violins}
% \end{figure*}

\begin{figure*}[!t]
     \centering
    \includegraphics[width=1\linewidth]{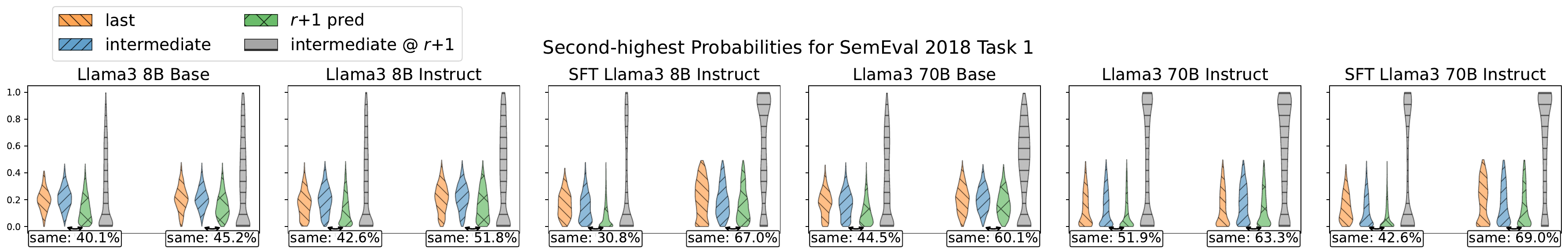}     
    \includegraphics[width=1\linewidth]{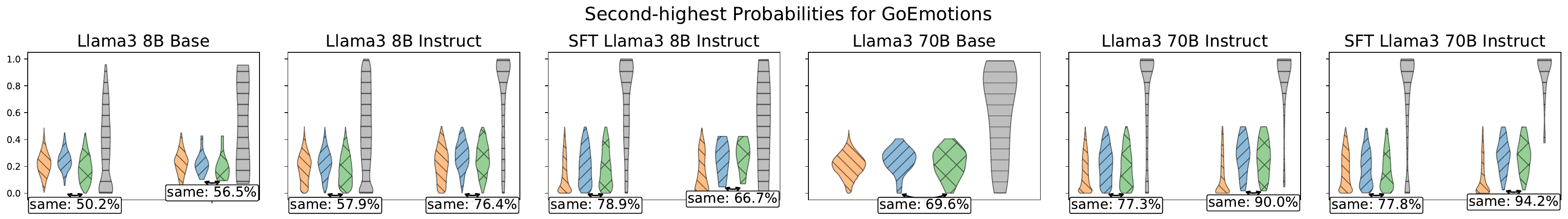}     
    \includegraphics[width=1\linewidth]{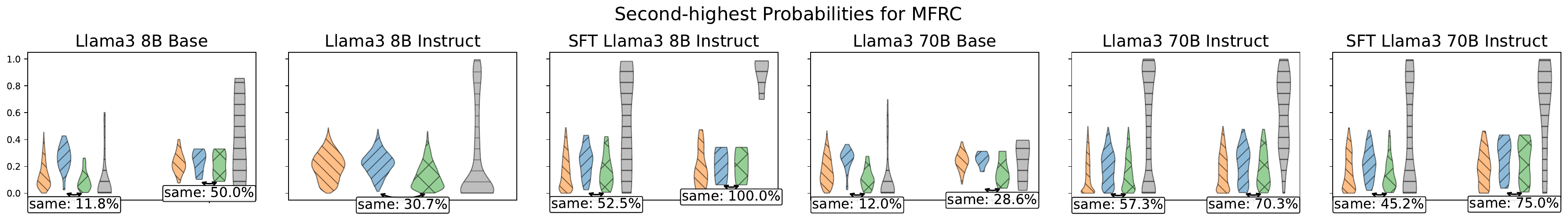}     
    \includegraphics[width=0.67\linewidth]{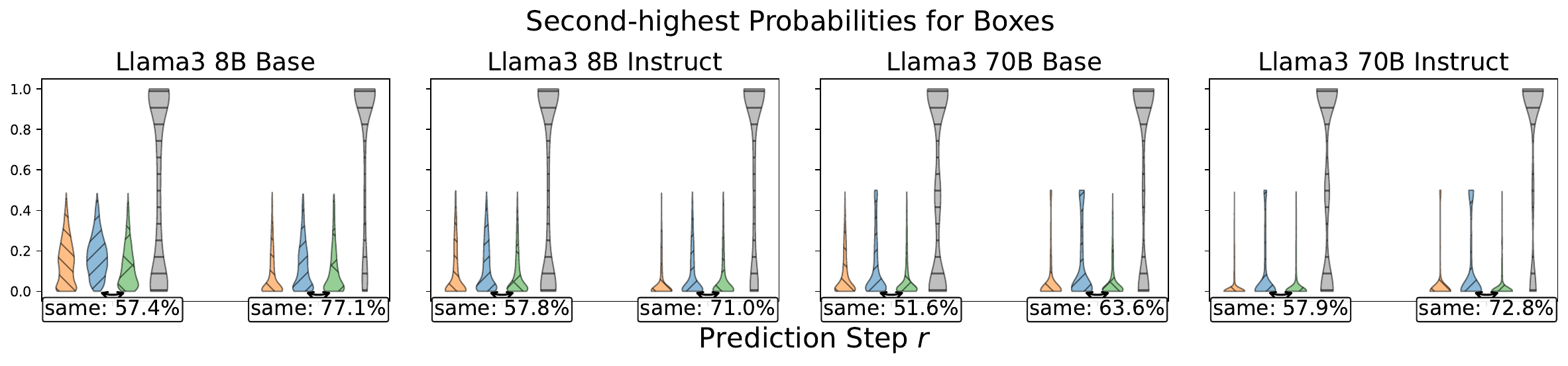}     

    \caption{Second-highest probabilities at each generation step when the \texttt{last} or an \texttt{intermediate} label is generated. We also show the probability at the current step of the label that is actually predicted in the next step ($r$+1 \texttt{pred}), the probability at the next generation step of the second highest probability of the current step (\texttt{intermediate @} $r$+1), and the percentage of cases the second-highest probability label at step $r$ and the prediction at $r$+1 is the \texttt{same}. LLM distributions show poor relative ranking, and little distinction between the \texttt{last} and \texttt{intermediate} settings. A single step only is shown when only up to labels were generated for all examples in a specific setting.}
    \label{fig:second-violins}
\end{figure*}

\label{sec:how_multi_label}

We evaluate whether LLMs produce diverse, consistent, and informative probability distributions. Specifically, we investigate whether the predicted probabilities at each generation step reflect the relative confidence of the LLM and whether the relative ordering of labels provides insight into future predictions. To this end, we analyze the distribution of the top two predicted probabilities at each label generation step, along with the entropy of the distribution, allowing us to assess how spiky the distributions are, that is, how close the top probability is to 1 and how low the entropy is. That is, we take the output probabilities of the model at each generation step where a label starts being predicted (if the LLM breaks a label into multiple tokens, then we take into account the probabilities for only the first token), extract the top two probabilities for further analysis, and also compute the entropy of the entire distribution.

We also compare the top probabilities to evaluate whether their relative values reflect the model’s confidence. Crucially, we examine the second-highest probability and track how it evolves in the subsequent generation step, and importantly how often the corresponding label is predicted next, as would be expected. By distinguishing between steps where the model continues generating more labels (denoted as \texttt{intermediate}) and steps where it predicts the final label (denoted as \texttt{last}), we assess whether the second-highest probability provides a meaningful signal about future behavior.% of the model. 

Finally, we test whether the relative order of the probabilities is informative by comparing the second-highest probability in the current generation step to that of the label generated in the next. That is to say, we look at the next generation step, see the label that was actually predicted, and then compare that label's probability in the current generation step compared to the probability of the second-highest label in the current step.

% \texttt{mid} and \texttt{last}

Figures~\ref{fig:top-violins} and \ref{fig:second-violins} show the results based on the predicted probabilities for all datasets using Llama3 8B and 70B Base, Instruct, and with Supervised Finetuning~\cite{ouyang2022training} (SFT; details in \S\ref{sec:appendix-models}). We show only up to the second step to avoid clutter. Corresponding entropy measures can be found in \S\ref{sec:appendix-entr}. We highlight key findings below.

\paragraph{Spikiness} We see that as the models become larger or are finetuned, the distributions start to concentrate around $100\%$. For instance, in \textbf{SemEval}, we see that Llama3 70B Instruct and SFT noticeably spike for both generation steps. In contrast, Llama3 8B Base has mode $\sim 40\%$. For \textbf{Boxes}, the objective benchmark, we observe even more pronounced spikes, with probability mass clustered around $\sim 100\%$ for all steps.

\paragraph{Sequential Spikiness} We observe that after the first label is generated, each additional label produced by the LLM is accompanied by a similarly spiky distribution centered on the newly predicted label. Interestingly, some distributions become spikier at later generation steps, potentially stemming from previously generated labels being assigned near-zero probability. % which constrains the candidate set.

\paragraph{Stopping Criterion} We find that models rarely have different distributions when predicting their last label compared to when they are going to continue predicting more labels, providing little to no indication of when they will stop predicting. Indeed, we would expect the distributions to resemble \textbf{MFRC} with the Base models, when the probabilities for the second highest labels are distinctly greater, the model continues to produce more labels. However, this distinction does not appear in most settings. For instance, \textbf{SemEval} has the same trends between both, and the second probabilities of some of the models are greater when the model stops generating (e.g., 70B Instruct and SFT), a counter-intuitive finding, because one would expect lower weight on the rest of the labels when the model would stop generating.

\begin{figure}[!tbp]
    \centering
    \includegraphics[width=1\columnwidth]{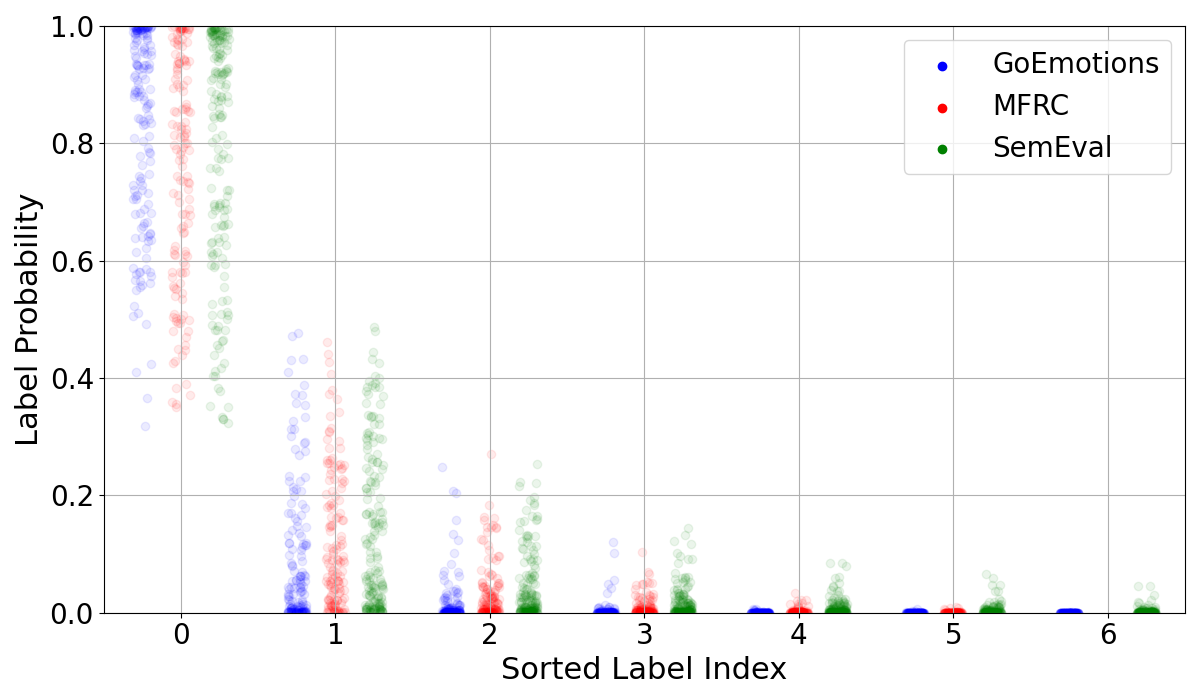}
    \caption{Sorted label probabilities when generating the first label for Llama3 70B Instruct. Most distributions are spiky, with the top label having near-1 probability.}
    \label{fig:spikiness}
\end{figure}

\paragraph{Relative Ranking} We demonstrate that LLMs do not reliably pick the second highest label as their next prediction, even if they continue predicting. For instance, in \textbf{SemEval}, the label with the second highest probability in the first step is not predicted next between $48.1\%$ and $69.2\%$ of the time across models. In \textbf{GoEmotions}, this behavior occurs between $22.2\%$ and $49.8\%$ of cases. In fact,
if we take the label with the second-highest probability in the current step $r$, and look at its probability in the next step $r$+1 (shown as \texttt{intermediate @} $r$+1), we see that it is clusters at 0. Similarly, when we look at the probability of the label predicted in step $r$+1, and see how its probability looked in the previous step $r$ (shown as $r$+1 \texttt{pred}), its probability tends to also be clustered around 0.
Notably, we find that if the second highest label at any step is not predicted as the next generated label, it will not be not predicted at all most of the time (see \S\ref{sec:appendix-2nd-not-predicted}). While is in some sense expected, since each generated label is newly conditioned on the previously generated labels (we verify this in \S\ref{sec:appendix-attn} by looking at the attention weights), it means that each generation step is only informative of the current label, since the \textit{relative ordering} of predicted labels is not predictive of subsequent behavior.

\paragraph{Language Modeling} From the previous two findings, we conclude that LLMs' distribution at the first (or any) generation step is not reflective of their confidence for each label, nor their subsequent behavior, suggesting \textbf{language modeling is interfering with classification}, causing the model to spike for every generation, an artifact of the autoregressive nature of LLMs, instead of generating a label distribution that is reflective of its confidence. We present more corroborating evidence in \S\ref{sec:results} with linear probing~\cite{hewitt2019designing}.

\begin{figure}[!ht]
    \includegraphics[width=0.95\columnwidth]{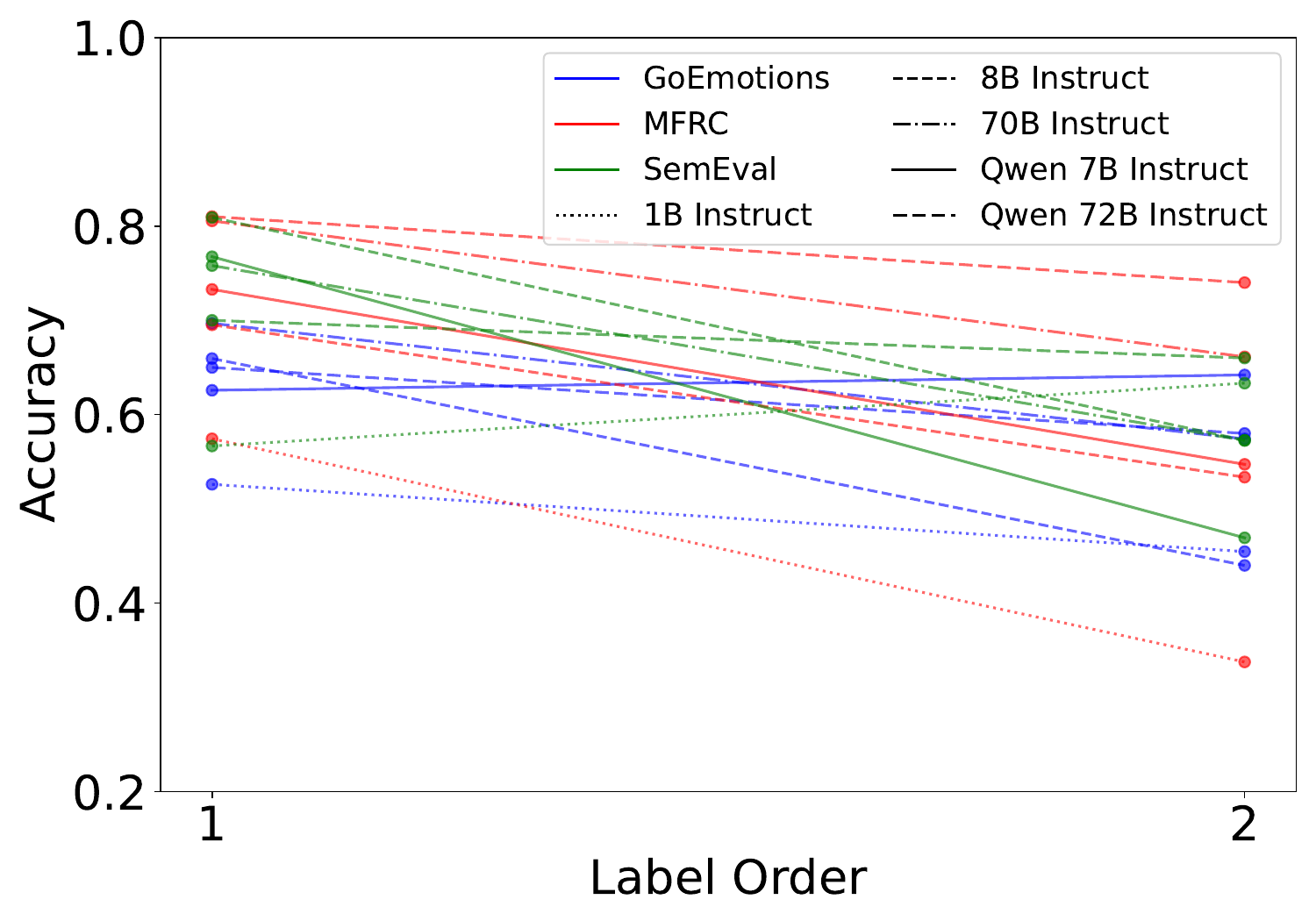}

    \caption{Average accuracy of the first and second label for multi-label generations based on the order in which it was generated, showing decreasing trends. Line color represents dataset and line pattern represents model size.}
    \label{fig:label_order_accuracies}
\end{figure}

\paragraph{Complete Distribution} We find that most label probability distributions are spiky, with the top label having probability near 1 and other labels sharply degenerating to near-0 probability even if later predicted (Figure~\ref{fig:spikiness}). We also find evidence that LLMs generate the most-likely label first, as the relative accuracy of each label drops between the first and second prediction in Figure~\ref{fig:label_order_accuracies}. Sequential spikiness explains these phenomena -- LLMs generate the most-likely label first with high confidence and do not consider what a less likely second label would be until the first label is fully generated. For the smaller models, we also observed a few instances where the model predicted the same label twice in a row.

% In Figure~\ref{fig:spikiness}, we present the complete distribution of all labels in the first generation step. Most label probability distributions are spiky, with the top label having probability near 1 and other probabilities sharply degenerating to near-0 probability. We also find that the relative accuracy of each label drops between the first and second prediction (Figure~\ref{fig:label_order_accuracies}). We hypothesize this is because LLMs generate the most-likely label first. This would also explain the spiky distribution, as the LLM does not need to model what the second label will be until the first label is fully generated. For the smaller models, we also observed a few instances where the model predicted the same label twice in a row.

\begin{figure}[!tbp]
    \includegraphics[width=\columnwidth]{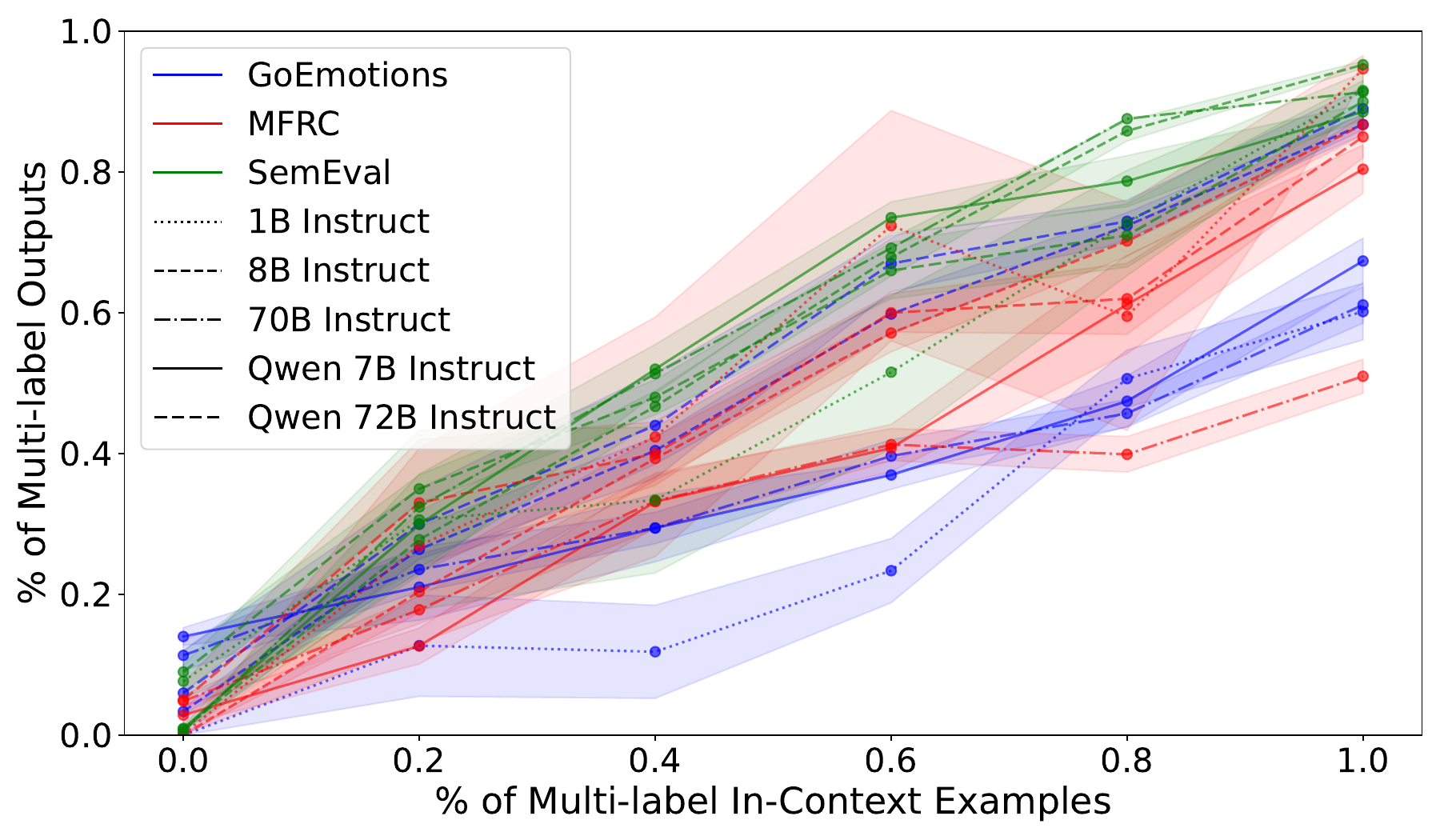}

    \caption{Percentage of outputs that are multi-label given the percentage of in-context examples that are multi-label in a 10-shot prompt. Line color represents dataset and line pattern represents model size.}
    \label{fig:multilabel_ICL}
\end{figure}

\paragraph{Rate of multiple predictions} Finally, we report that the label type of in-context prompts greatly influences the rate of multi-label output. We show in Figure~\ref{fig:multilabel_ICL} how the percentage of multi-label (as opposed to single or no label) examples roughly corresponds to the percentage of multi-label output across models and datasets. Learning to predict multi-label outputs must be highlighted very clearly in the in-context examples, suggesting that single-label formats have dominated the training of the model. Overall, these analyses show that LLMs do not create well-calibrated distributions when generating multiple labels; instead, they generate spiky distributions, classifying labels one at a time.

\paragraph{Generalizability} To ensure our findings generalize to other model families, we replicate the main results for the Qwen 2.5~\cite{team2024qwen2} family of models in \S\ref{sec:appendix-qwen}, showing identical results to the Llama family. Moreover, we experimented with an LLM with multiple decoding heads, Medusa~\cite{pmlr-v235-cai24b}. Given its ability to predict multiple tokens at a time, the aforementioned behaviors might not be present in such models. Contrary to this assumption, we show in \S\ref{sec:appendix-medusa} that the model behaves in identical ways. Finally, in \S\ref{sec:appendix-label-order} we examine whether the label order in the instructions has a role in these phenomena, finding strong effects.
\section{Multi-Label Distribution Alignment}
\label{sec:distribution_alignment}

\begin{table*}
\centering
\begin{adjustbox}{width=0.96\textwidth}
    \begin{tabular}{cccccccccccccc}
    \midrule
     & & \multicolumn{6}{c}{Single-Label Datasets}& \multicolumn{6}{c}{Multi-Label Datasets} \\
\cmidrule(lr){3-8} \cmidrule(lr){9-14}
     &   & \multicolumn{3}{c}{Hatexplain} & \multicolumn{3}{c}{MSPPodcast} & \multicolumn{3}{c}{GoEmotions} & \multicolumn{3}{c}{MFRC} \\
\cmidrule(lr){3-5} \cmidrule(lr){6-8} \cmidrule(lr){9-11} \cmidrule(lr){12-14}
    & &  NLL $\downarrow$ & L1 $\downarrow$ & F1 $\uparrow$ & NLL $\downarrow$ & L1 $\downarrow$ & F1 $\uparrow$ & NLL $\downarrow$ & L1 $\downarrow$ & F1 $\uparrow$ & NLL $\downarrow$ & L1 $\downarrow$ & F1 $\uparrow$ \\ \midrule \midrule
    \multirow{2}{*}{\rotatebox{90}{\resizebox{0.9cm}{!}{Baseline}}} & Compare-to-None & 1.66 & 0.81 & 0.58 & 2.63 & 1.37 & 0.29 & 23.93 & 4.71 & 0.27 & 5.34 & 1.85 & 0.51 \\
     & Hard Predictions & 9.86 & 0.90 & 0.58 & 13.65 & 1.47 & 0.30 & 24.11 & 1.31 & 0.39 & 19.70 & 1.07 & 0.59 \\

    \midrule
    \multirow{3}{*}{\rotatebox{90}{\resizebox{1.4cm}{!}{Test-Time}}} & Unary Breakdown & \textbf{0.91} & 0.94 & 0.47 & \textbf{1.55} & 1.45 & 0.30 & 3.60 & 1.32 & 0.43 & 2.49 & 1.27 & 0.51 \\
     & Binary Breakdown & 1.12 & 1.06 & 0.29 & 1.65 & 1.44 & 0.24 & 7.62 & 2.64 & 0.41 & 3.55 & 2.11 & 0.41 \\
     & Max-Over-Generations & \na & \na & \na & \na & \na & \na & 4.04 & 1.27 & 0.39 & \textbf{2.32} & 0.92 & 0.63 \\

    \midrule
    \multirow{4}{*}{\rotatebox{90}{\resizebox{1.5cm}{!}{Supervised}}} & BERT & 2.69 & \textbf{0.73} & \textbf{0.66} & 4.29 & \textbf{1.27} & \textbf{0.38} & 2.72 & \textbf{0.63} & \textbf{0.64} & 3.00 & \textbf{0.43} & \textbf{0.82} \\
     & Linear Probing & \skipped & \skipped & \skipped & \skipped & \skipped & \skipped & \textbf{2.42} & 0.71 & 0.56 & 2.81 & 0.44 & 0.81 \\
     & SFT Outputs & \skipped & \skipped & \skipped & \skipped & \skipped & \skipped & 14.76 & 0.80 & 0.58 & 10.45 & 0.57 & 0.69 \\
     & SFT Max-Over-Generations & \na & \na & \na & \na & \na & \na & 4.15 & 0.72 & 0.57 & 4.87 & 0.54 & 0.73 \\

    \midrule
\end{tabular}
\end{adjustbox}
    \caption{Distribution alignment scores for Llama3 70B Instruct on single and multi-label datasets between LLM and human distributions. $\textbf{F1}\uparrow$ is the example-F1 score. \na: Not applicable to single-label setting. \skipped: Not supplied to avoid clutter, and due to environmental considerations, since single-label settings are not our focus.
    % Supervised methods are finetuned on Llama3 70B Instruct (except BERT) on a held-out train set.
    }
    \label{tab:llama_3.3_70B_results}
\end{table*}

\begin{figure*}[!ht]
\centering
    \includegraphics[width=1\linewidth]{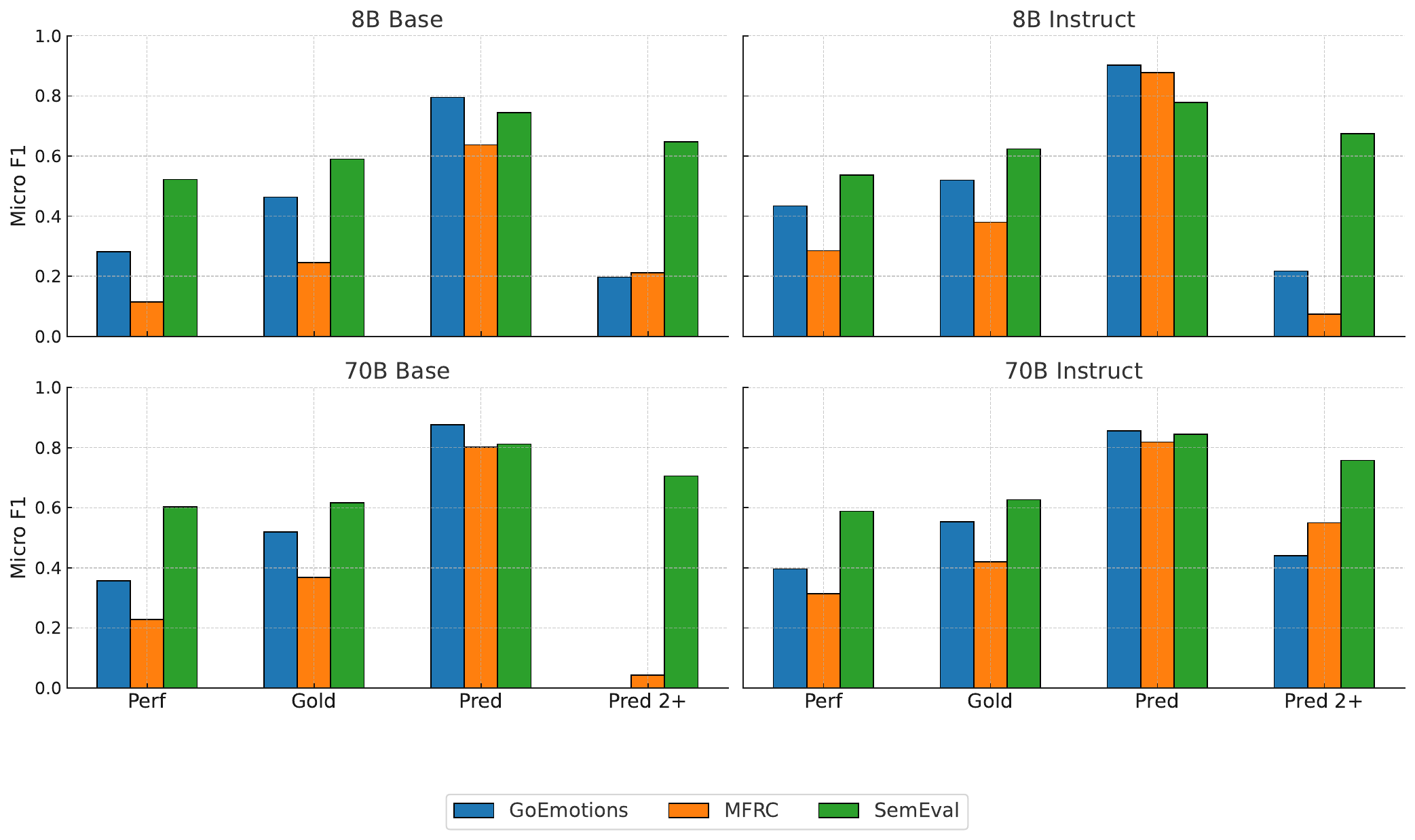}
    \caption{\textbf{Micro F1} $\uparrow$ of linear probes trained and evaluated on gold labels (\textit{Gold}), trained and evaluated on model predictions (\textit{Pred}), and evaluated on predictions beyond the first generated label (\textit{Pred 2+}). For comparison, we also show the performance of the model (\textit{Perf}). Embeddings are from the last layer for the first generated label.}
    \label{fig:lin_probe}
\end{figure*}

To test how interpretable and calibrated the LLM-derived distributions are, we propose \textbf{multi-label distributional alignment} as a core task. Our focus in this work is multi-label subjective tasks, because they allow degrees of belief, and so allow us to evaluate model confidence, not just predictions, in multi-label settings.

\subsection{Task Formulation for Multi-Label}
In the single-label setup, a probability distribution is produced over a label set $L$. However, in the multi-label case, each example can have an arbitrary number of labels, each of which has its own binary probability of appearing (in practice, labels are additionally correlated). Thus, multi-label distributions are $|L|$ binary probabilities.

\subsubsection{Human Distribution Estimation}
Our underlying assumption is that given a task with subjective labels and multiple interpretations, the ``truth'' of the label is better represented as a confidence distribution over a potential label set. In this interpretation, for data point $d$, an annotation represents a single sample $a \sim H(d)$, where $H$ is the underlying human distribution. Then, denoting $\mathbb{I}$ as the indicator function, for label $l \in L$, we approximate our empirical human-annotation distribution using annotator set $A$ as:
\vspace{-6px}
\begin{equation}
    \hat{H}_l(d; A) = \frac{1}{|A|} \sum_{a_i\in A} \mathbb{I}[l\in a_{i}(d)].
\end{equation}
\vspace{-6px}

\subsubsection{Distribution Alignment Metrics}
\label{subsec:metrics}

% We compute two metrics per example to measure how well the human annotator sample fits a given LLM probability distribution, negative log likelihood (NLL) and L1 distance, and one performance metric\footnote{\S\ref{sec:appendix-caveat} for a note on the need for a performance metric}, example-F1, an F1-variant designed for multi-label settings \cite{giraldo-forero_2015}. 
We compute the negative log likelihood (NLL), L1 distance, and example-F1~\cite{du_2019} to evaluate how well the empirical distribution aligns with the LLM-derived distribution. Example-F1 is a variant of F1 that can be evaluated per example.

\paragraph{NLL}
Conceptually, NLL measures if a distribution is confidently wrong about any answer. Given a discrete probability distribution $Q_d$ and a set of labels $G_d = \{g_i|\ i\in [m], g_i\in L\}$, we compute the likelihood of $G_d$ as $\prod_{g\in G_d}^m P_{Q_d}(g)$, where $P_{Q_d}(l_i)$ is the probability of $l_i$ under $Q_d$.
Taking the negative logarithm gives NLL. The best distribution that explains a sample minimizes NLL.

\paragraph{L1 Distance}
One shortcoming of NLL is that it disproportionately penalizes small differences near 0, e.g., penalizing a likelihood of $10^{-7}$ much more than $10^{-2}$, despite their practical similarity. L1 distance solves this problem by comparing the absolute difference of each label probability to its frequency in the sample: $\sum_{l\in L}|P_{Q_d}(l) - \hat{H}_l(d; A)|$. L1 distance measures if the general shape of the distributions match
% \begin{align*}
%     \text{L1 distance}_D = \sum_{l}^L \bigg{|}P_D(l) - \frac{\text{count}_S(l)}{|S|}\bigg{|}
% \end{align*}

% \paragraph{Example-F1}
% Example-F1 is a variant of the standard F1 score that can be evaluated per-example in a multi-label setting; following \citet{du_2019}, we prefer it over Micro-F1 because we want to give equal weight to an example that contains no labels and an example that contains many labels. For a given example $i$ with True Positives (labels correctly identified), False Positives (wrong labels classified), and False Negatives (labels not identified), we have:
% \begin{equation}
%     \text{Example-F1}_i = \frac{2 \cdot \text{TP}_i}{2 \cdot \text{TP}_i + \text{FP}_i + \text{FN}_i}
% \end{equation}

\subsection{LLM Distribution Methods}
\label{subsec:llm_methods}

To investigate the task of distribution alignment in the multi-label setting, we propose methods which are categorized into three groups: \textit{baseline methods}, \textit{test-time methods}, and \textit{supervised methods}.

\subsubsection{Baseline Methods}

\paragraph{Compare-to-None}

We use the output distribution of the labels at the point at which the model generates its first label token (excluding, for example, formatting tokens).
%Namely, for each label, we extract its first token (note that a label can be broken into multiple subwords). Then, we look for these tokens in the generated text of the model. Finally, we collect the language modeling logits for the earliest of these tokens.
However, the individual values of raw logits hold little interpretability as their value is only meaningful in the context of the rest of the tokens. We propose to compare the logit score of each label to the logit score of the ``none'' label to get an estimate of how likely that label is to occur independent of the other logits, leveraging the null prediction to contextualize the value of the logits. Let $S(l_i)$ be the logit score for label $l_i$; we can therefore determine the \textbf{logit score difference} for each label $d_i = S(l_i) - S(l_\text{none})$.
We then apply the sigmoid function to $d_i$ for a valid probability: $P(l_i=1|d_i) = \sigma(d_i)$. % We also experimented with calibrating the sigmoid function into a scaled logistic regression model (i.e., $\sigma(w \cdot d_i + b)$ on the dev set), but found that this did not improve performance, so we use the normal sigmoid function.

% While this seems like a reasonable baseline, research has shown that contemporary LLMs are not well-calibrated after RLHF~\cite{ouyangTrainingLanguageModels2022}. In fact, this work showed that LLMs predict the next token with very high confidence, leading to a skewed distribution for the different labels, failing to provide us with a more nuanced understanding of the relationship between different labels.

% Multiple additional confounding factors exist in the multi-label case. First and foremost, due to the causal nature of the LLMs, the model does not generate predictions from the joint distribution of the labels, but takes a conditional approach. For instance, the second highest probability when the first label is predicted might not accurately reflect the labels that the model is subsequently going to produce, as the next label will be generated conditioned on the first. Moreover, arbitrary ordering effects might influence the predictions of the model. For example, if the labels have been presented in an alphabetical order, then the model would be incentivized to arbitrarily increase its weight for the label with the lower alphabetical order, but that may not reflect its confidence for the label

\paragraph{Hard (Actual) Predictions}
We take the labels that the model actually outputs autoregressively; we set these values to $1-\epsilon$ and otherwise $\epsilon$ to avoid arithmetic issues with NLL.

\subsubsection{Test-Time Methods}

% \begin{figure*}
%     \includegraphics[width=\textwidth]{figs/methods_sketch.jpg}

%     \caption{[USING SKETCH RIGHT NOW, WILL MAKE REAL FIGURE LATER] Overview of the three new zero-shot methods for multi-label probability distribution approximation introduced in this work: Max-Over-Generations, Unary Breakdown, and Binary Breakdown.}
%     \label{fig:methods}
% \end{figure*}

Findings from \citet{niculescu-mizil_2005} indicate that binary tasks are generally well-calibrated. Even though modern LLMs are very different from the basic neural networks tested in this paper, we were inspired to design several different approaches that ``break down'' multi-label classification into smaller steps. For these methods, we investigated Monte Carlo sampling methods but found this approach simply added noise over directly calculating the label probabilities analytically.

\paragraph{Unary Breakdown: Label-wise Preference}

In this approach, we create a binary classification problem for each individual label, similar to the approach taken by \citet{li_2020}. Namely, for a given example, we create a prompt that includes the original document to be classified, but instead we present a single label and query the model if the label is ``reasonable''. We directly extract the probabilities for the ``reasonable'' label, which conforms to the independence property of multi-label probabilities, because each label can be assigned a value $\in[0, 1]$ without constraints or normalization. $|L|$ runs (one per label) per document are required.

\paragraph{Binary Breakdown: Pair-wise Preference}

We break down a single example into multiple binary comparisons between all label pairs ($\binom{|L| + 1}{2}$ runs per example), and then leverage the outcomes of these comparisons to derive probabilities for the labels. Namely, for every pair of labels, we provide both labels to the model and ask the model to select one of them as better representing the input. We derive the probabilities for the two labels by applying softmax on the two logits. We then use the Bradley-Terry model~\cite{bradley1952rank} to rank the labels based on their pairwise performance. Specifically, to estimate logit scores $S$ with pairwise probabilities that label $l_i$ is better than $l_j$, we have ${P(l_i \text{ is better than } l_j) = p_{i>j} = \sigma(s_i - s_j)}$, where $\sigma$ is the sigmoid function.
This is calculated by minimizing the predictive loss $\mathcal{L}$:
\begin{equation}
\begin{split}
    \mathcal{L} = -\frac{1}{2}& (\sum_{i, j} p_{i>j} \cdot \log(\sigma(s_i - s_j)) \\& + (1 - p_{i>j}) \cdot \log(\sigma(s_j - s_i))).
\end{split}
\end{equation}
In order to calculate the multi-label probabilities, similar to \textbf{compare-to-none}, we introduce a ``none'' label into the label set and derive final probabilities by comparing the Bradley-Terry logit scores of a given label to the ``none'' logit score. We also consider using strict 1's and 0's instead of probabilities, similar to ELO ranking \cite{elo_1978} in \S\ref{appendix_binary_probs_outcomes}, but find using probabilities to be more performant.

\paragraph{Max-Over-Generations} We take the probability distributions for \textbf{every} label generation step, and the final probability for each label is equal to the maximum value achieved over all distributions. This approach is a soft version of the \textbf{Hard Predictions} baseline, and requires access to model scores.

\subsubsection{Supervised Methods}

We compare our approach with three supervised methods: \textbf{Finetuned BERT}, \textbf{Linear probes}~\cite{hewitt2019designing} on the first label token of the last layer, and \textbf{SFT}, all described in \S\ref{sec:appendix-models}. We also use \textbf{Linear probes} for interpretability purposes~\cite{li2021implicit} to study the informational content of the models' embeddings.

\subsection{Experimental Setup}

We apply our methods on the same Llama models (see \S\ref{sec:appendix-models}).
We test our proposed approaches on the main test set (details in \S\ref{sec:appendix-splits}). We test on the multi-label datasets of \textbf{GoEmotions} and \textbf{MFRC} that contain individual annotator labels. We also include evaluation on two single-label subjective datasets (details in \S\ref{sec:appendix-single-dataset}), \textbf{HateXplain}~\cite{mathew2021hatexplain} and \textbf{MSP-Podcast}~\cite{Lotfian_2019_3} to contextualize our multi-label findings.
\subsection{Results}
\label{sec:results}

\paragraph{Distribution Alignment} We report distribution alignment results in Table~\ref{tab:llama_3.3_70B_results} for Llama3 70B (results for 8B in \S\ref{sec:appendix-8b-alignment}). Overall, we find that Test-Time and Supervised methods outperform both baseline methods. We draw particular attention to the max-over-generations method, which significantly outperforms both baselines with little additional computational overhead other than storing model scores across multiple generation steps. We see that unary breakdown performs similarly well to max-over-generations, as isolating each label's validity independently disentangles the bias of language modeling from the classification task. As a downside, unary breakdown incurs $|L|$ times the generations per example. Surprisingly, we find that BERT performs the best of the supervised methods, which we use as additional evidence that LLMs classify labels one at a time, not simultaneously. 

\paragraph{Linear Probing} The linear probing method ranks as the second best baseline, so the hidden states during first-label generation alone seem, at first glance, to contain enough information to perform well on the tasks. However, in Figure~\ref{fig:lin_probe}, we present a more detailed analysis with linear probes. In addition to model and probing performance, we present the probes' capability of predicting the predictions of the model themselves (i.e., the probes are trained on the predictions). We present the performance on the predictions on the \textit{Pred} column, showing, as expected, much higher performance. However, when we look at how well the probes can predict any label after the first (\textit{Pred 2+}), we see a substantial degradation in performance. Note that the task in theory becomes easier as we remove a label from the problem. This degradation suggests that linear probing performs well mostly due to its high accuracy of the first label and has less predictive power for any future labels, which aligns with our findings that LLMs predict labels one at a time. Even after supervised training, embeddings of the first label generation do not contain enough information to predict any subsequent labels.

% the good performance of linear probing is derived in large part from predicting the first label correctly. This directly shows that the embeddings also do not contain enough information for the next labels, suggesting they are not appropriate for multi-label classification. Nonetheless, we see improvements on the larger Instruct model.

\paragraph{Effect of Instruction Tuning} In \S\ref{sec:appendix-ft-alignment}, we demonstrate that finetuned models generally achieve higher performance, yet their NLL is worse. This result supports previous findings that finetuned model are more confident, since NLL punishes confidently wrong predictions more.
\section{Conclusion}
\label{sec:conclusion}

We provide the first account of how LLMs perform multi-label classification and find that LLMs generate spiky probability distributions and appear to predict labels one at a time rather than jointly. 
We argue that language modeling interferes with multi-label classification, making it difficult to interpret model confidences for labels until they are predicted. We provide supportive experimental evidence, demonstrating that a full generation of output is required to analyze LLMs' label confidences, and highlight the inconsistencies in the label probabilities across generation steps. Finally, we formulate the task of distribution alignment in the multi-label setting and propose novel methods and baselines to estimate better multi-label distributions from language models. We conclude that much work is still required in order to create distributions from LLMs that match the human distribution in responses to subjective language tasks.
\section{Limitations}
\label{sec:limitations}
There are several potential limitations in this work. First, our assumption of underlying empirical distributions derived from human annotator samples relies on the fact that the annotators are in fact valid and representative samples of the underlying true distribution. % of the human distribution. 
This does not account for the possibility that different annotators may be biased in the same way and that combining their annotations does not remove this bias. Additionally, we limit our analysis to the Llama model family, which is inherently constrained to these models' specific training and finetuning regimens. We acknowledge the possibility that our insights into multi-label generation for LLMs may differ for different model families. Finally, our proposed methodologies of unary and binary breakdowns also increase the computational cost when compared to a single label generation, and that while these methods may show improvement over single generations, this increased cost is certainly a limitation towards their adoption.
\section*{Acknowledgments}

This project was supported in part by funds from NSF CIVIC, and USC-Capital One Center for Responsible AI Decision Making in Finance. The authors thank Thanathai Lertpetchpun, Kleanthis Avramidis, Emily Zhou and Jihwan Lee for helpful comments.

\bibliography{citations}
\clearpage

\appendix

\section{Additional Implementation Details}
\label{sec:appendix-implementation_details}

\subsection{Label Probabilities}

Throughout \S\ref{subsec:llm_methods}, we generate softmax probabilities of the label set by constraining the logit scores to just those of the initial tokens of labels. This deviates slightly from the true label probabilities, as we ignore all non-label token values during the softmax; however we note that, in practice, the softmax probabilities over just the label set do not deviate much from their probabilities over the entire vocabulary set, as the majority of top logits are label tokens.

\subsection{Multi-Label Datasets}

\paragraph{GoEmotions} The seven emotion ``clusters'' are: \textit{admiration} (includes pride, gratitude, relief, approval, realization), \textit{anger} (includes disgust, annoyance, disapproval), \textit{fear} (includes nervousness), \textit{joy} (includes amusement, excitement, love), \textit{optimism} (includes desire, caring), \textit{sadness} (includes remorse, embarrassment, disappointment, grief), and \textit{surprise} (includes confusion, curiosity). The clustering was performed using the hierarchical clustering algorithm, applied on the correlations between emotions, as described in~\cite{demszky2020goemotions}.

\paragraph{MFRC} The six moral foundations are: \textit{care}, \textit{proportionality}, \textit{equality}, \textit{purity}, \textit{authority}, and \textit{loyalty}.

\paragraph{SemEval} The eleven emotion labels are: \textit{anger}, \textit{anticipation}, \textit{disgust}, \textit{fear}, \textit{joy}, \textit{love}, \textit{optimism}, \textit{pessimism}, \textit{sadness}, \textit{surprise}, and \textit{trust}.

\subsection{Single-Label Datasets} \label{sec:appendix-single-dataset}

% \paragraph{MMLU (objective)~\cite{wang_2024}} Multiple choice benchmark covering 57 subjects such as algebra and chemistry, varying from elementary to advanced difficulty. Up to 10 potential answers are provided for each example, with as little as 3 for some. Contains 12k samples.

\paragraph{HateXplain~\cite{mathew2021hatexplain}} Benchmark of hateful and offensive speech. Each document is labeled as \textit{offensive}, \textit{hateful}, or \textit{normal}, and where necessary it also contains the target of that sentiment. 
% Contains 15k train, 1922 development, and 1924 test examples.
Each sample was assigned to 3 annotators.

\paragraph{MSP-Podcast v1.11~\cite{Lotfian_2019_3}} Utterances from podcasts that have been labeled for emotion. The dataset comes with ground truth transcriptions, which we leverage to perform language modeling.
% Contains 84k examples for training, 20k for development, and three separate test sets of 30k, 15k and 2k examples each.
5.3 annotators on average were assigned to each sample.

\subsection{Dataset splits} \label{sec:appendix-splits}

For Figures~\ref{fig:top-violins} and \ref{fig:second-violins}, we perform inference on the Base and Instruct models on the entire training set to get the largest population of data points we can. However, for the SFT models, since we needed a large enough training set, we use the train split to finetune the model and perform inference on the dev and test sets.

For the linear probes, we train on the train set and evaluate on the dev and test sets.

For the rest of our experiments, and for each dataset, we create two testing sets: a "multi-label only" set, containing data that exclusively has multiple ground truth labels, which we use in \S\ref{sec:how_multi_label}; and a main testing set, which contains a uniform number of data across three label types (no label, single label, and multi-label) and annotator disagreements (no disagreement and has disagreement) for our experiments in \S\ref{sec:distribution_alignment}. For each test set we select 200 data points per dataset due to exploding number of runs we require for the methods we propose (e.g., unary requires a run per label). In the prompt, half of the in-context examples contain multiple labels.

\subsection{Models} \label{sec:appendix-models}

We use the following models, all downloaded from HuggingFace and implemented in PyTorch:
\begin{itemize}
    \item Llama3 1B Instruct\\ (\texttt{meta-llama/Llama-3.2-1B-Instruct})
    \item Llama3 8B Base\\ (\texttt{meta-llama/Llama-3.1-8B})
    \item Llama3 8B Instruct\\ (\texttt{meta-llama/Llama-3.1-8B-Instruct})
    \item Llama3 70B Base\\ (\texttt{meta-llama/Llama-3.1-70B})
    \item Llama3 70B Instruct\\ (\texttt{meta-llama/llama-3.3-70B-Instruct})
\end{itemize}
We used NVIDIA A100 80GB VRAM GPUs for 70B models, and NVIDIA A40 for smaller models.

\paragraph{SFT}

Our supervised finetuning pipeline simply involves prompting an LLM with the same instructions and prompt template as the other models, but without the 10 demonstrations that we otherwise use. We used LoRA~\cite{hu2022lora}. During inference, because we noticed a tendency for the model to respond with differing formats, we still used a 10-shot format to standardize the output.

\paragraph{Unary breakdown}

We specifically use the term "reasonable" given the subjective nature of the tasks where multiple labels may be appropriate, as we found that using "yes" or "no" directly sometimes causes the model to assign a more appropriate label even if both labels are applicable.

\paragraph{BERT}

For the BERT results, we have used Demux~\cite{chochlakisLeveragingLabelCorrelations2023}. We use the same training regime as in the original paper, using the intra loss with a coefficient of $0.2$ for the multi-label settings, but training only on the train set instead of integrating the dev set in training after early stopping. For the single-label settings, we simply switch to using the cross-entropy loss instead of the binary cross-entropy.

\paragraph{Linear Probes}

We derive the hidden state at the last layer of the first \textit{label} token that the model generates. We normalize and downsample with a factor of 4 using truncated SVD (to accommodate for the smaller dataset size compared to the hidden state dimension, especially of the 70B models). We then train one logistic regression model per label using \texttt{scikit-learn}'s \texttt{Logistic Regression}.
\subsection{Caveat on NLL and L1} \label{sec:appendix-caveat}

In the multi-label setting, since every possible label has the potential to be included in an example, each sample technically contains data on every label, with the majority of labels being set to 0 (i.e., not assigned to the example). In scenarios where the majority of labels are 0, a degenerate solution of a "fixed" distribution, where all values are set to a constant such as 0.1, often performs very well. Thus, it is important to evaluate  pure alignment metrics such as NLL and L1 in conjunction with performance metrics such as accuracy or F1, as fixed distributions will perform very poorly on those metrics.
\section{Prompt Examples}
\label{appendix:prompt_examples}

\begin{table}[!ht]
    \centering
    \footnotesize
    \renewcommand{\arraystretch}{1}
    \begin{tabular}[t]{@{}p{23em} @{}}

    \textbf{Prompt examples} \\
    \midrule

    \textbf{Regular prompt (GoEmotions)} \\
    \narrowbotc{Classify the following inputs into none, one, or multiple the following emotions per input: joy, optimism, admiration, surprise, fear, sadness and anger. Output exactly these emotions and no others.\\\\Input: "Can I speak to the Suns' manager?"\\\{"label": ["surprise"]\}\\\\Input: Problem is that picture is there before you ever set foot in the cabin; [NAME] put it there. Just makes it creepy.\\} \\ \\

    \textbf{Unary Breakdown Prompt Example (Hatexplain)} \\
    \narrowbotc{Classify the following question-label pairs as either "reasonable" or "unreasonable". Output either "reasonable" or "unreasonable" and nothing else.\\\\Question: that or only date asians either way the white race is dead\\Label: offensive\\Is the label reasonable: unreasonable\\\\Question: i want to hate you but i can not\\Label: normal\\Is the label reasonable:} \\ \\

    \textbf{Binary Breakdown Prompt Example (MFRC)} \\
    \narrowbotc{Classify the following question into one of the two following labels. Output either "a" or "b" depending on which label is better and nothing else.\\\\Question: It’s just capitalism is a pure distillation of those bad traits\\a. none\\b. loyalty\\Correct Label: b\\\\Question: His abuse will get worse. He's literally already accusing you, what will he do if he doesn't believe you're being faithful?\\a. proportionality\\b. loyalty\\Correct Label: } \\\\

    % \textbf{} \\
    % \narrowbotc{} \\\\
    
    \end{tabular}
    \caption{Example prompts}
    \label{tab:prompt-example}
\end{table}

We present some prototypical prompts we use throughout our whole study in Table~\ref{tab:prompt-example}.
\section{Binary Breakdown with Probabilities vs. Outcomes}
\label{appendix_binary_probs_outcomes}

\begin{figure*}[!htb]
     \centering
     \begin{subfigure}[b]{0.49\textwidth}
         \centering
         \includegraphics[width=\textwidth]{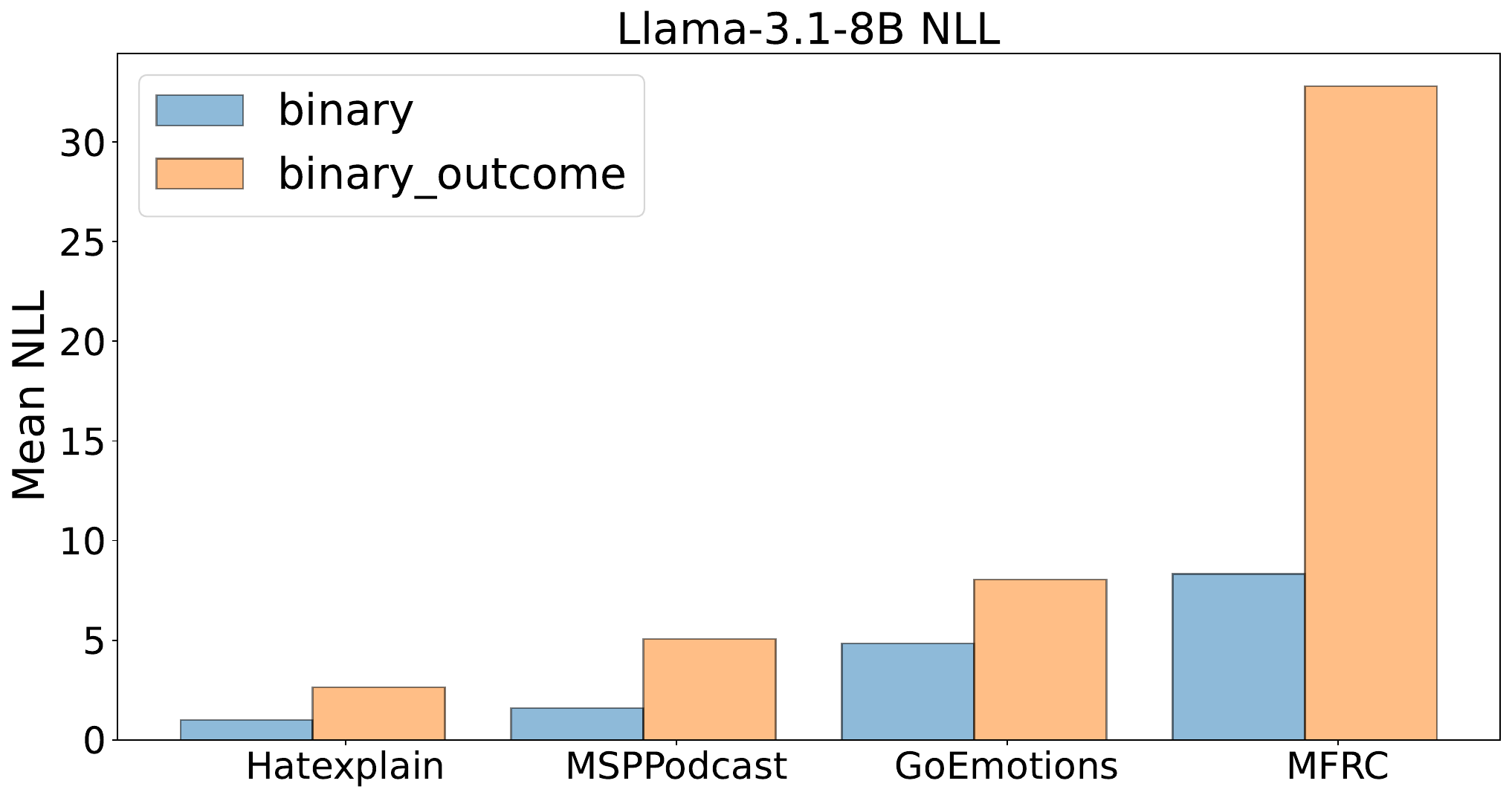}
     \end{subfigure}
     \hfill
     \begin{subfigure}[b]{0.49\textwidth}
         \centering
         \includegraphics[width=\textwidth]{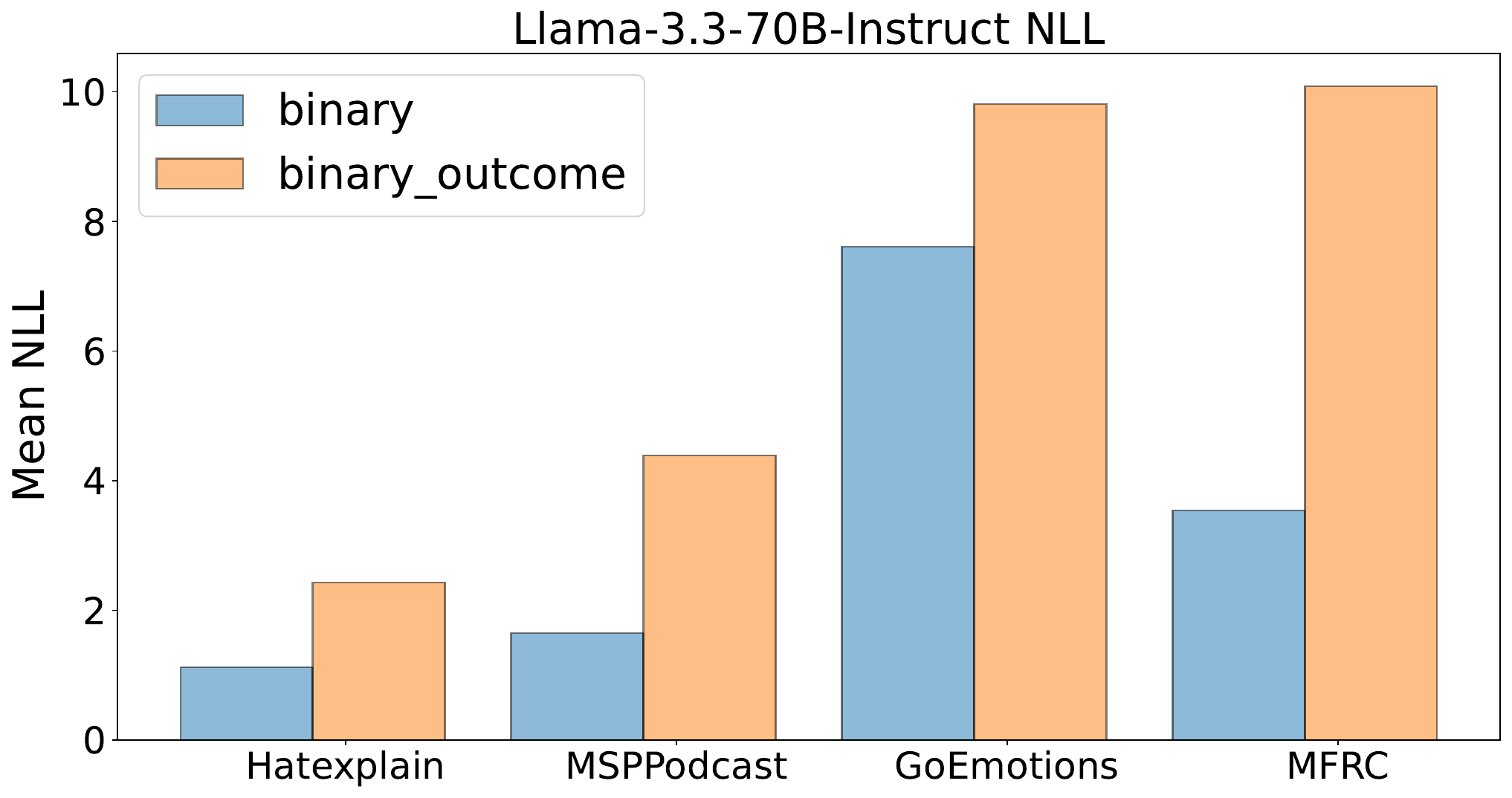}
     \end{subfigure}
     \hfill
     \begin{subfigure}[b]{0.49\textwidth}
         \centering
         \includegraphics[width=\textwidth]{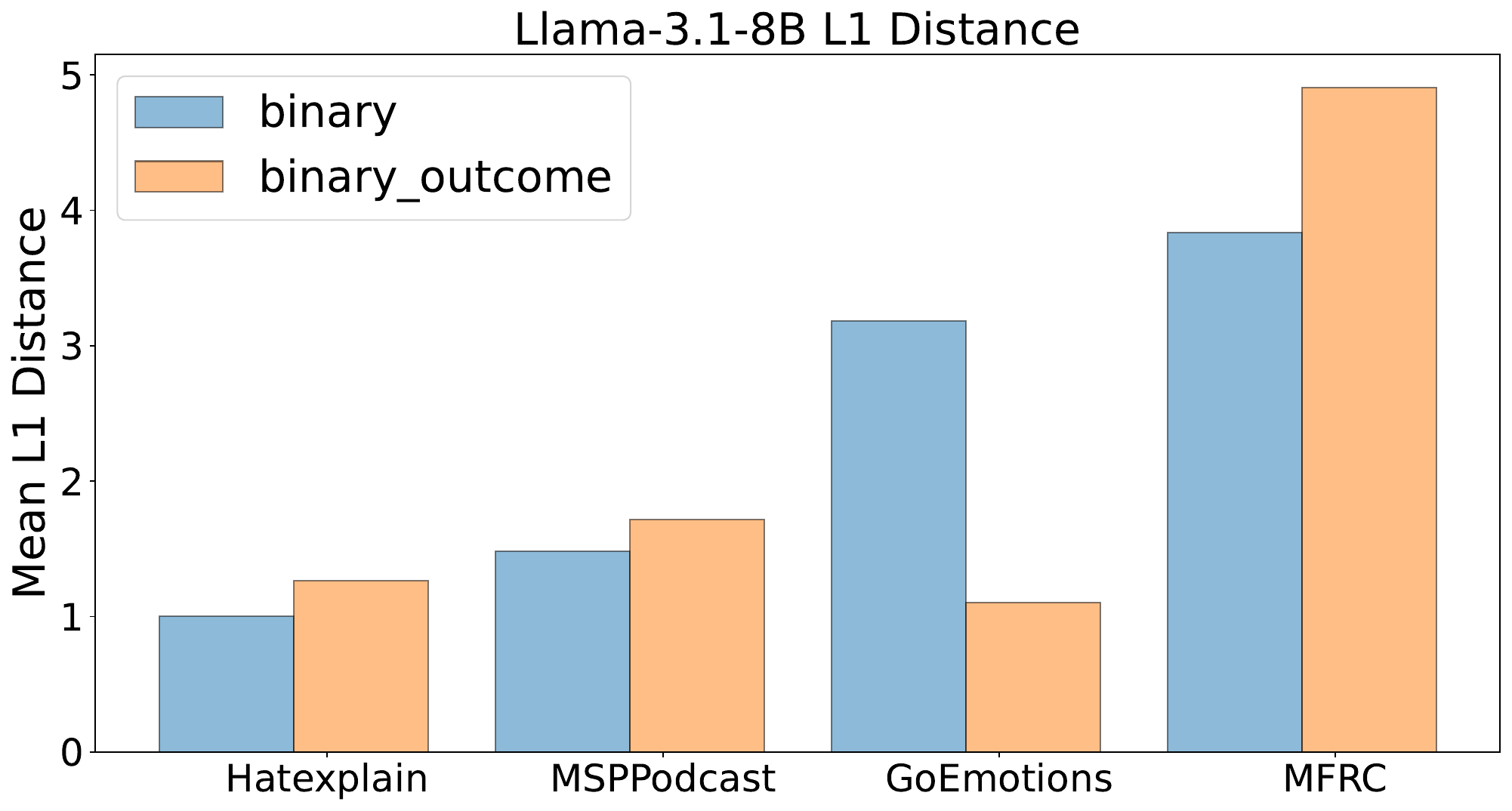}
     \end{subfigure}
     \hfill
     \begin{subfigure}[b]{0.49\textwidth}
         \centering
         \includegraphics[width=\textwidth]{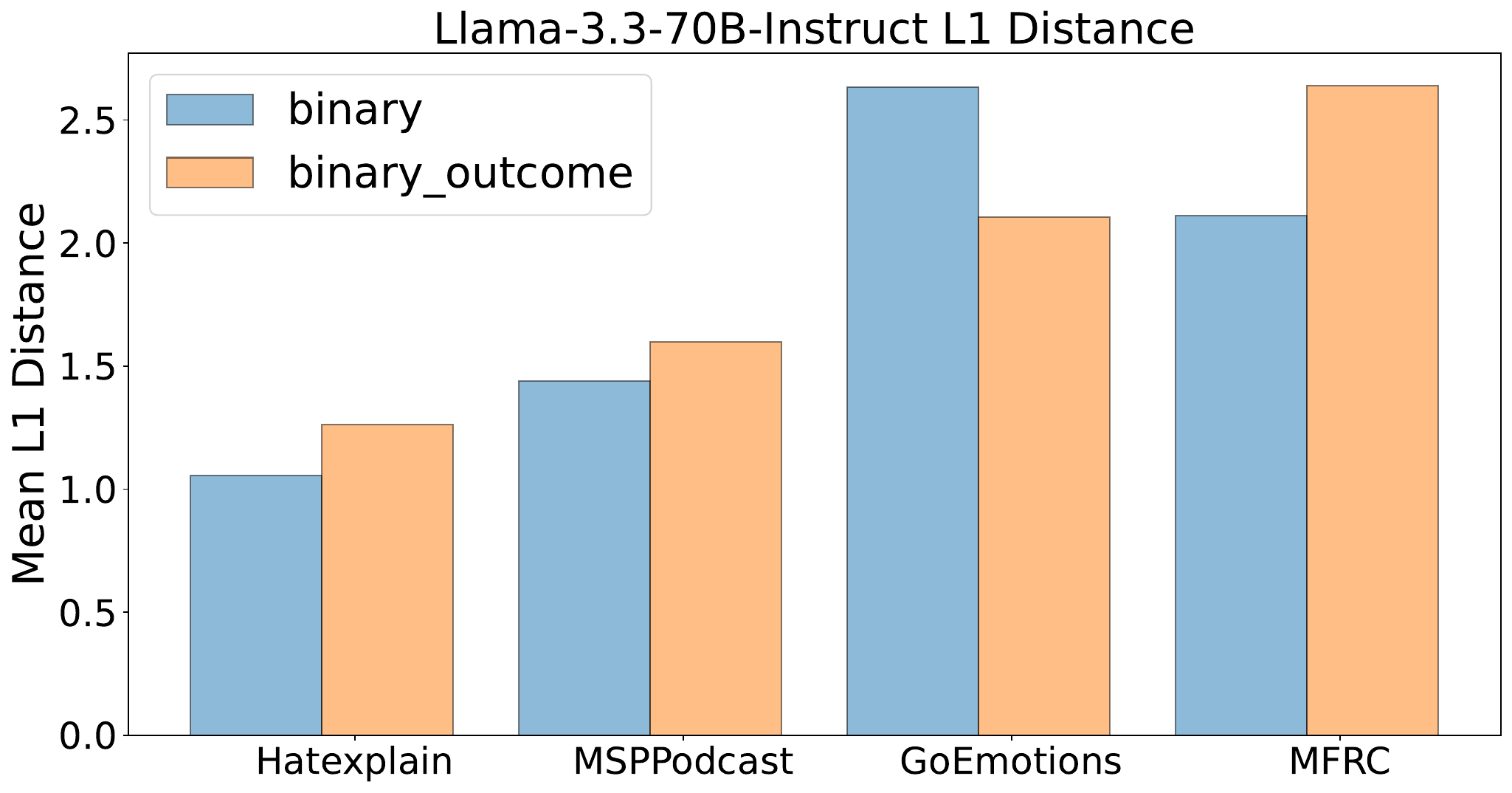}
     \end{subfigure}
     \hfill
     \begin{subfigure}[b]{0.49\textwidth}
         \centering
         \includegraphics[width=\textwidth]{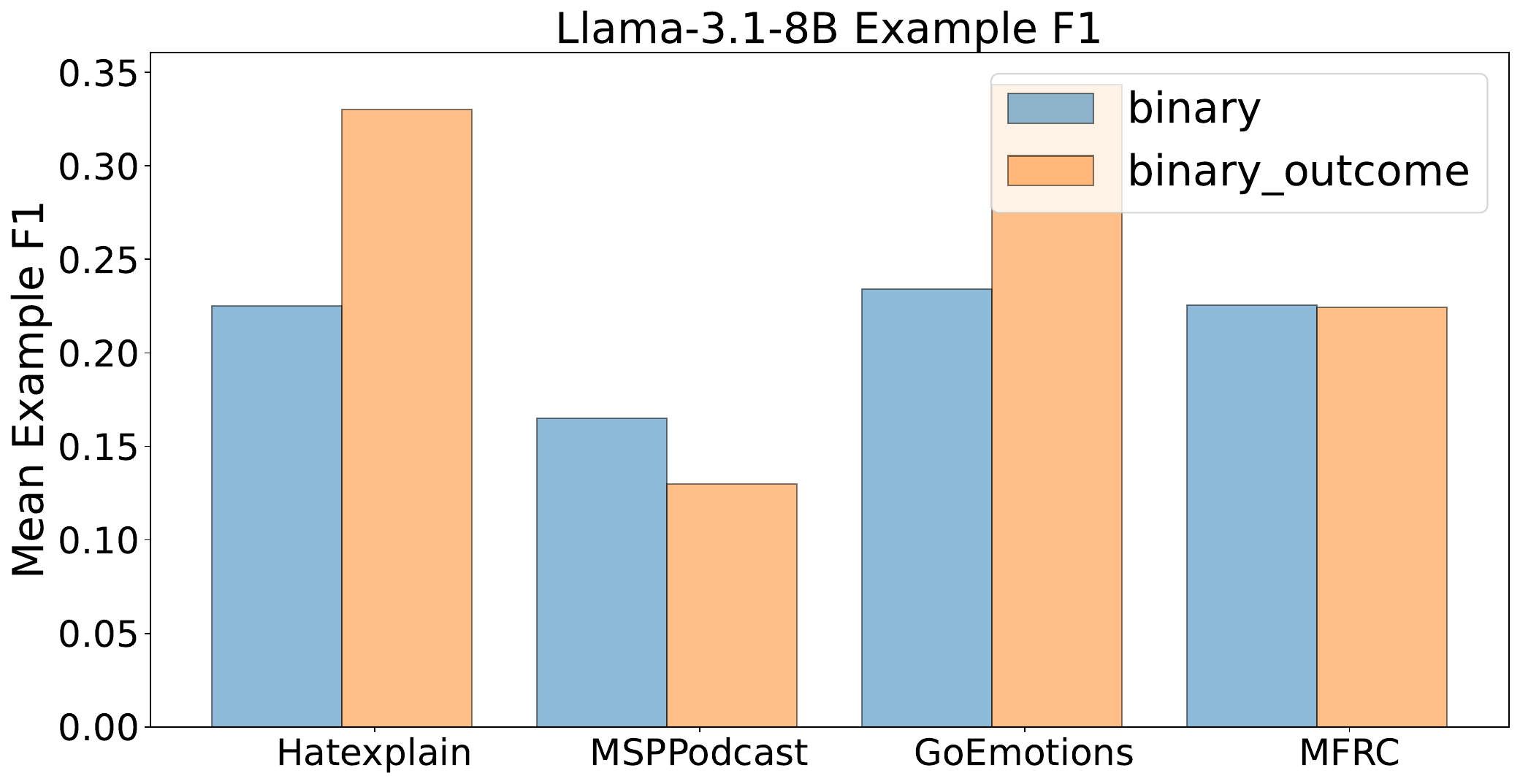}
    \end{subfigure}
     \hfill
     \begin{subfigure}[b]{0.49\textwidth}
         \centering
         \includegraphics[width=\textwidth]{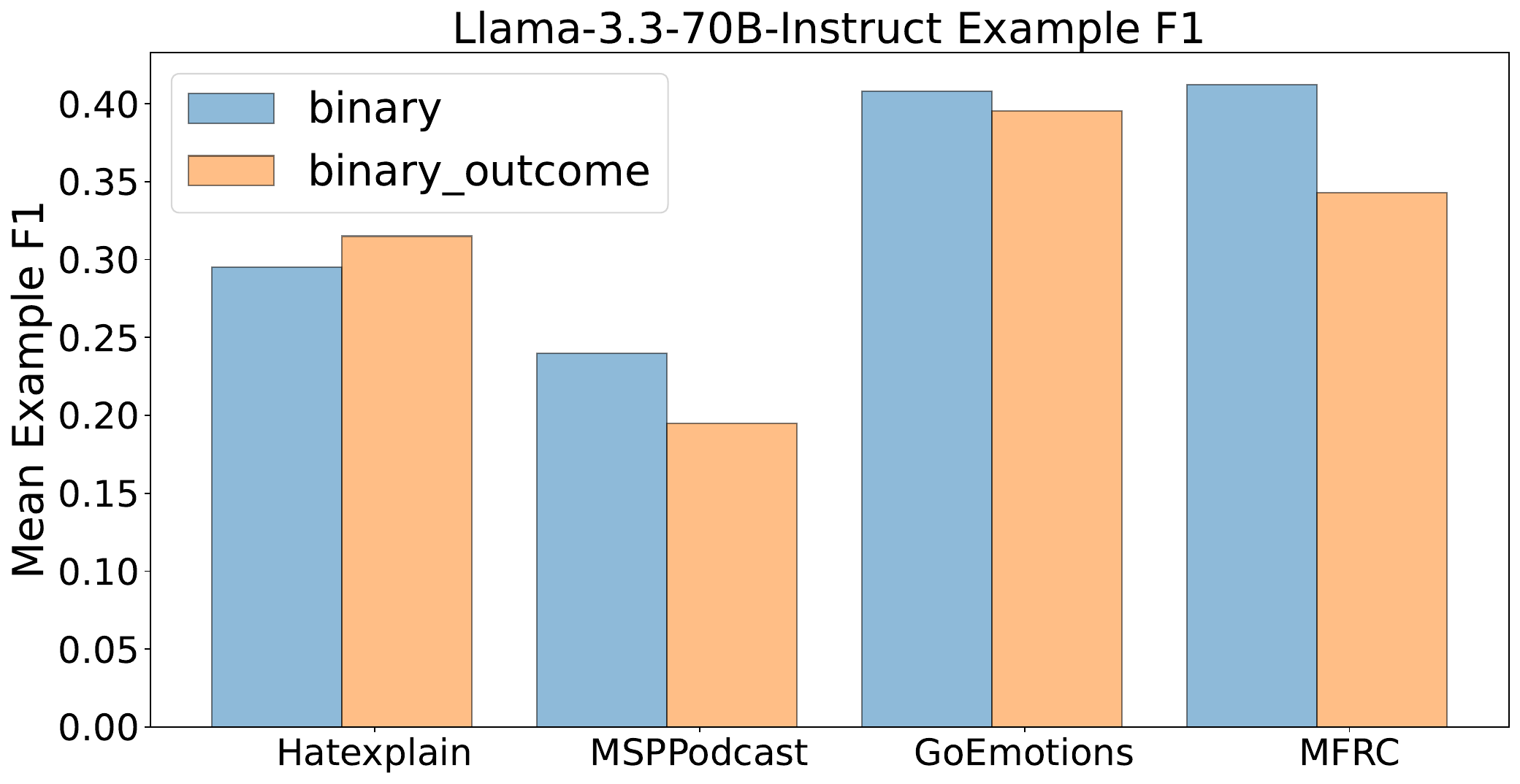}
     \end{subfigure}
     \caption{Comparison of binary breakdown when using the pairwise probabilities (``binary'') versus using pairwise outcomes (``binary\_outcome'', i.e. rounding probabilities to 0 and 1).}
     \label{fig:pairwise-outcomes}
     
\end{figure*}

We investigate two methods of binary breakdown in Figure~\ref{fig:pairwise-outcomes}: using the soft probabilities between 0 and 1 (i.e., $p \in [0, 1]$ for preferring one label to the other and using hard outcomes (i.e., $p \in \{0, 1\}$). We find that using for L1 distance and F1, the preferred approach varies between datasets, but for NLL, using probabilities is always preferred. We find that when a single label is dominant, meaning it is preferred to every other label, using probabilities calibrates the breakdown better than using hard outcomes, as dominant labels still never achieve 100\% probability in their comparisons. We therefore conclude that using binary breakdown with probabilities rather than outcomes is the better approach.

\section{Additional Results on LLM Multilabel Capabilities} \label{sec:app-extra-results}

\subsection{Probabilities: Alternative view}

\begin{figure*}[!htb]
     \centering
    \includegraphics[width=1\linewidth]{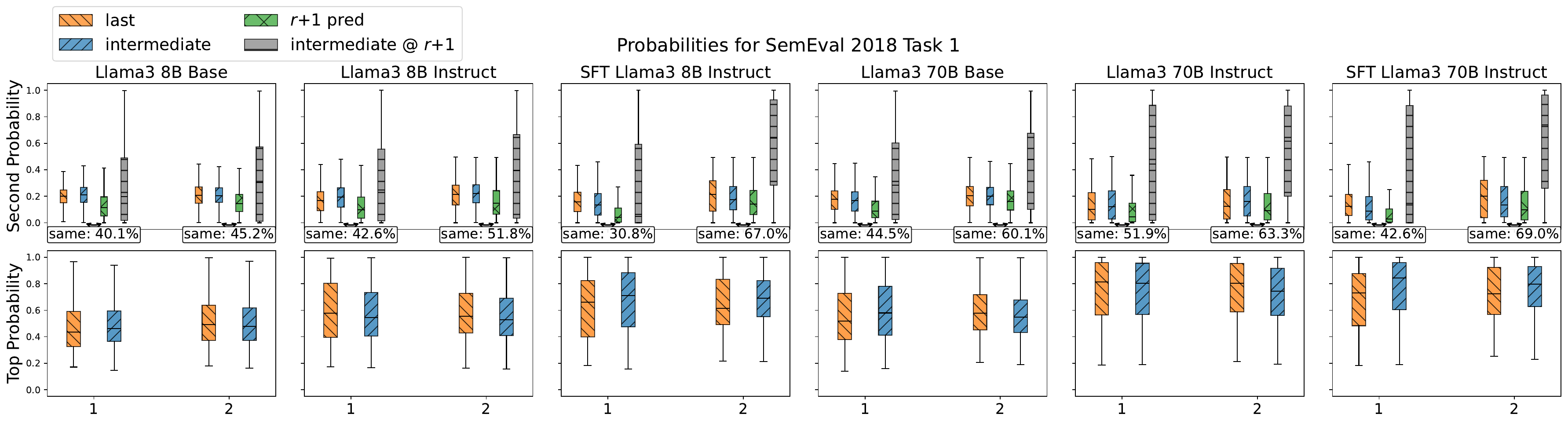}     
    \includegraphics[width=1\linewidth]{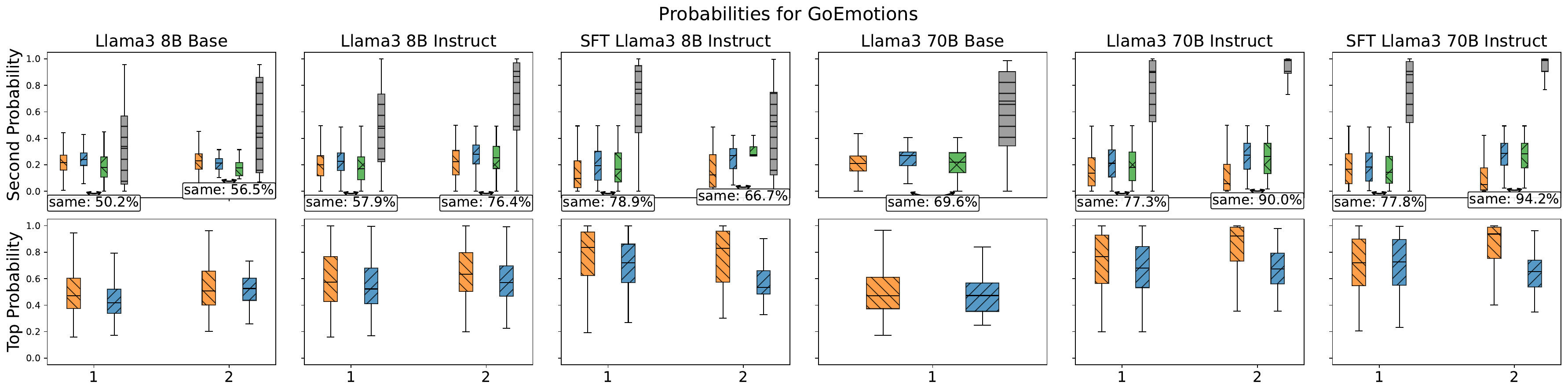}     
    \includegraphics[width=1\linewidth]{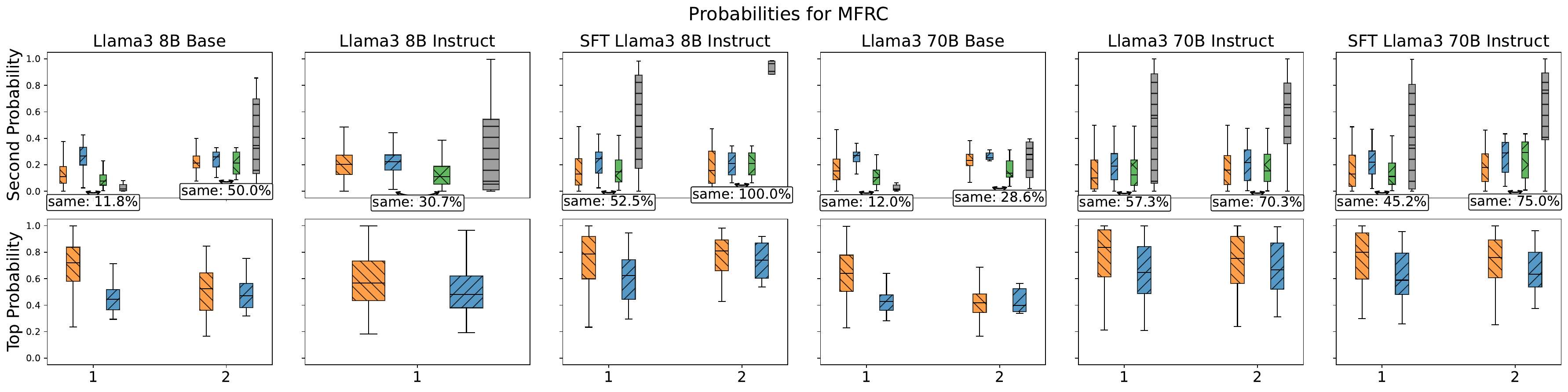}     
    \includegraphics[width=0.67\linewidth]{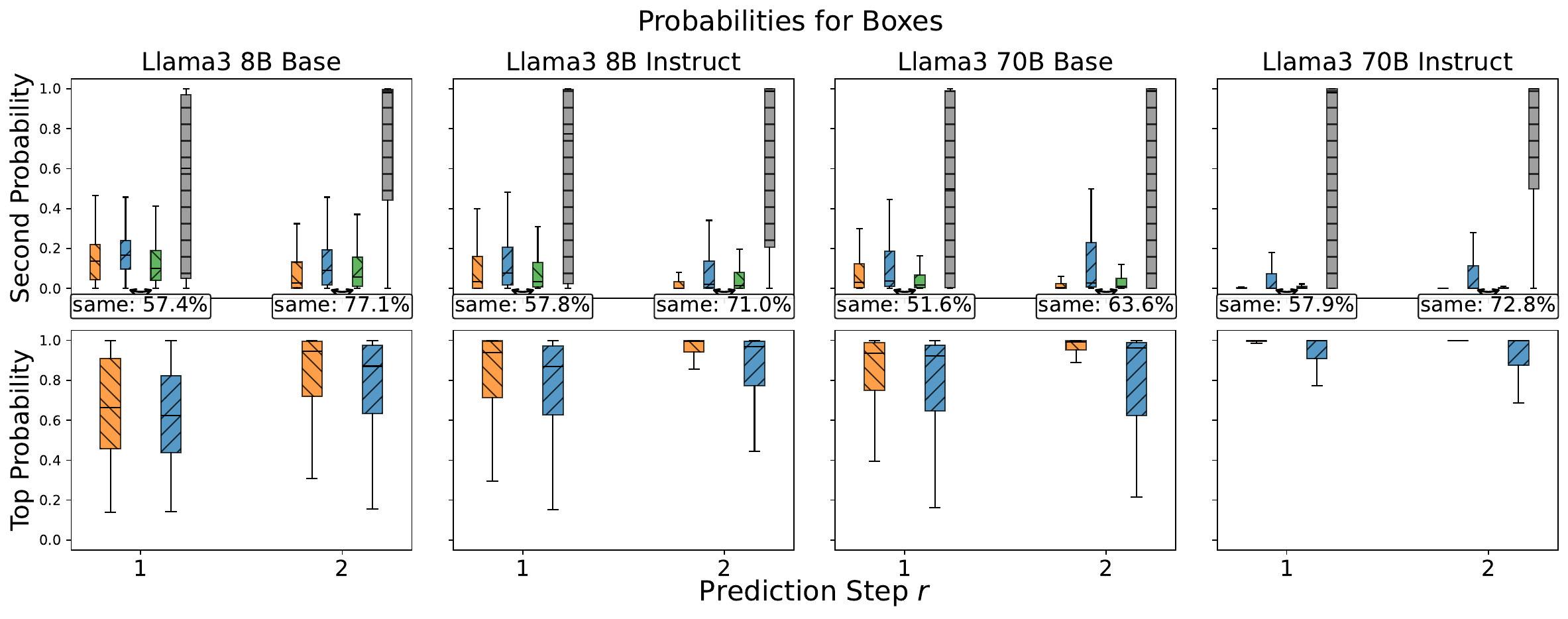}     

    \caption{Top two probabilities at each generation step $r$ (up to two for brevity) when the \texttt{last} label is generated, or when a \texttt{intermediate} label is generated. Shown are for four datasets, one per row. In each row, the bottom subfigure shows the top probability, and the top the second highest probability, in addition to the probability of the label that was actually predicted next at the current step ($r$+1 \texttt{pred}), and the probability at the next generation step of the second highest probability (\texttt{intermediate @} $r$+1). Also shown is the percentage of cases the second-highest probability label at $r$ and the prediction at $r$+1 were the \texttt{same}. A single step only is shown when only up to labels were generated for all examples in a specific setting.}
    \label{fig:main-boxplots}
\end{figure*}

For completeness, in Figure~\ref{fig:main-boxplots} we also present the equivalent box plots of Figures~\ref{fig:top-violins} and \ref{fig:second-violins}.

\subsection{Entropy of Predictions} \label{sec:appendix-entr}
\begin{figure*}[!htb]
     \centering
    \includegraphics[width=1\linewidth]{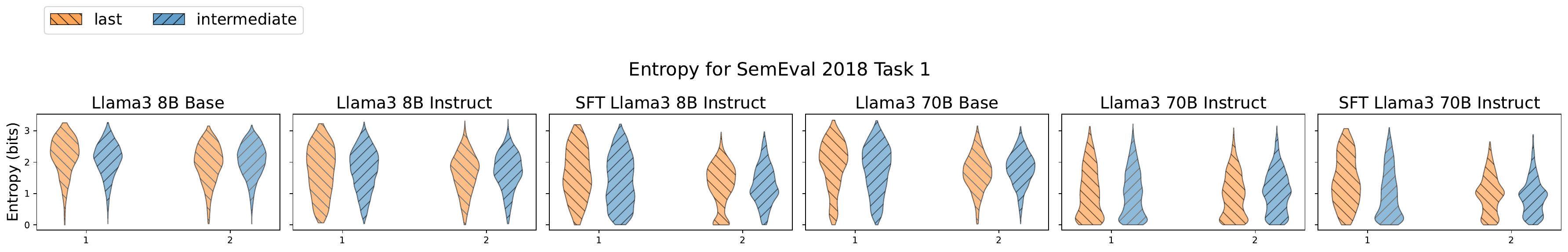}     
    \includegraphics[width=1\linewidth]{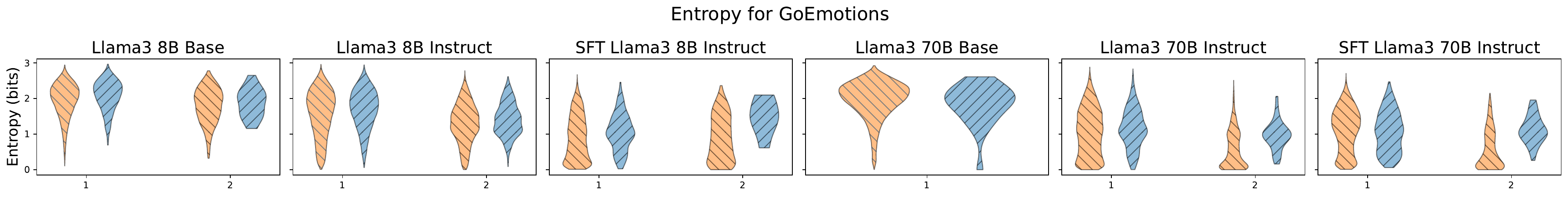}     
    \includegraphics[width=1\linewidth]{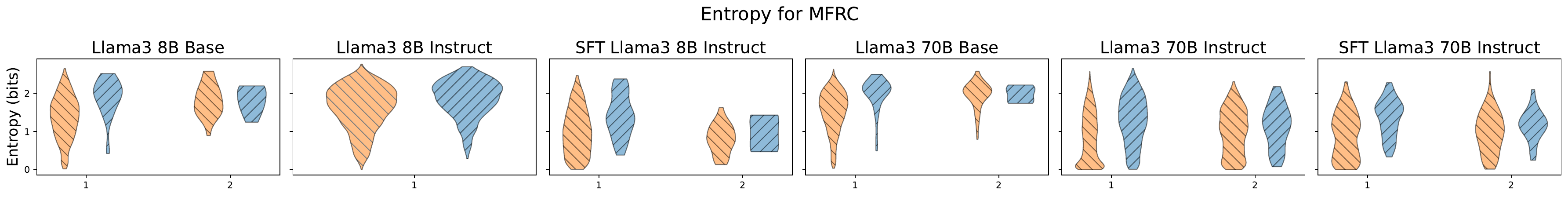}     
    \includegraphics[width=0.5\linewidth]{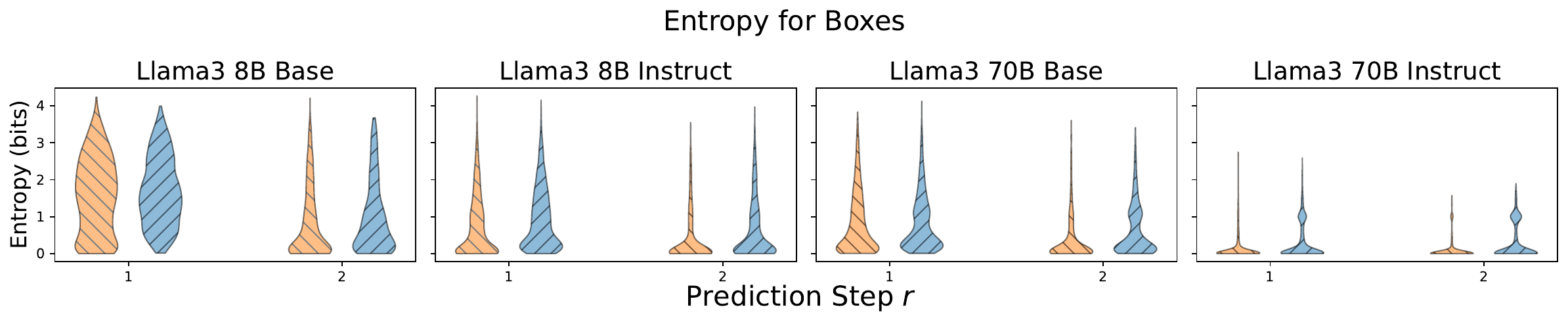}     

    \caption{Entropies of prediction distributions at each generation step $r$ when the \texttt{last} label is generated, or when a \texttt{intermediate} label is generated, shown for the first two label generation steps. A single step only is shown when only up to labels were generated for all examples in a specific setting.}
    \label{fig:entropies}
\end{figure*}

We also present the entropies of the predictions in Figure~\ref{fig:entropies}. Again, for all datasets but for \textbf{MFRC}, we see that the trends are indistinguishable between when the model will generate more labels compared to when it predicts its last label, showing little evidence for properly calibrated probability distributions on multi-label tasks. 

\subsection{Inconsistencies in second highest label scores} \label{sec:appendix-2nd-not-predicted}

In this section, we report the probability that the label associated with the second highest probability at any given generation step is, in fact, never predicted by the model if not predicted in the immediate next step. We limit our evaluation only to steps where the model does continue to predict more labels afterward, skipping the instances where the model stops predicting. In Table~\ref{tab:not-predicted}, we see that the label does not appear in the predictions at least $78.4\%$ of the time in \textbf{SemEval}, $91.3\%$ in \textbf{GoEmotions}, $89.9\%$ in \textbf{MFRC}, and $56.8\%$ in \textbf{Boxes}. Note that, as shown in Figure~\ref{fig:second-violins}, the second ranked label is not predicted immediately after a large percentage of time, resulting overall in large inconsistencies in the probabilities and the predictions of LLMs.

\begin{table*}[!htb]
    \centering
    \begin{tabular}{lcccccc}
        \toprule
        & \textbf{8B Base} & \textbf{8B Instruct} & \textbf{8B SFT} & \textbf{70B Base} & \textbf{70B Instruct} & \textbf{70B SFT}  \\
        \cmidrule(r){2-7}
        \multicolumn{1}{l|}{\textbf{SemEval}} & 88.1 & 85.3 & 90.4 & 78.4 & 78.8 & 82.8 \\
        \multicolumn{1}{l|}{\textbf{GoEmotions}} & 99.3 & 95.4 & 91.3 & 92.9 & 93.4 & 96.7 \\
        \multicolumn{1}{l|}{\textbf{MFRC}} & 100 & 99.7 & 94.7 & 94.0 & 96.4 & 89.9 \\
        \multicolumn{1}{l|}{\textbf{Boxes}} & 86.1 & 70.8 & - & 72.4 & 56.8 & - \\
        \bottomrule
    \end{tabular}
    \caption{Percentage \% of cases the second highest label in probability was not predicted at all at any subsequent step when it was not predicted immediately afterward, despite the model predicting at least one more label.}
    \label{tab:not-predicted}
\end{table*}

\begin{figure}[!htb]
    \centering
    \includegraphics[width=1\columnwidth]{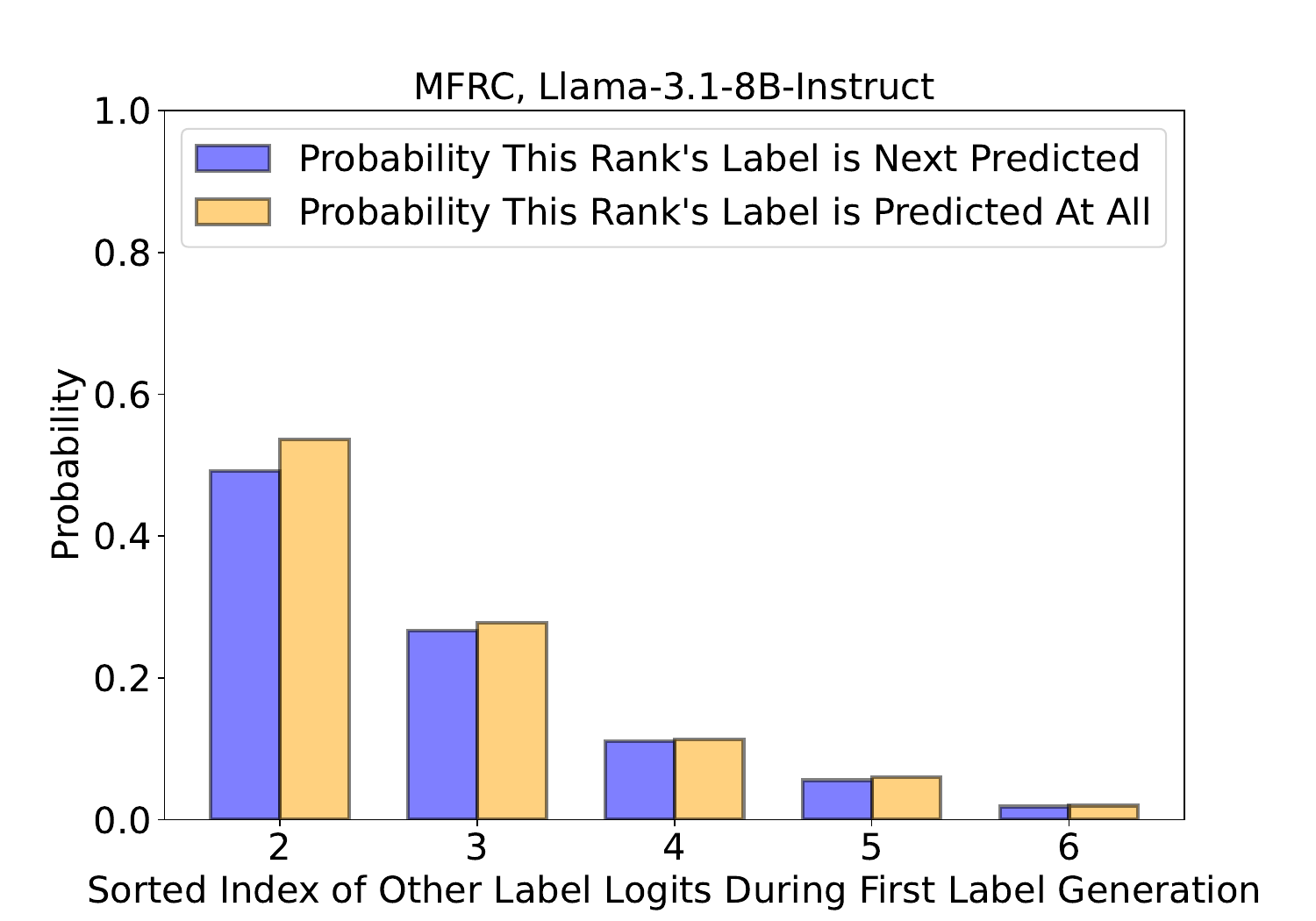}
\includegraphics[width=1\columnwidth]{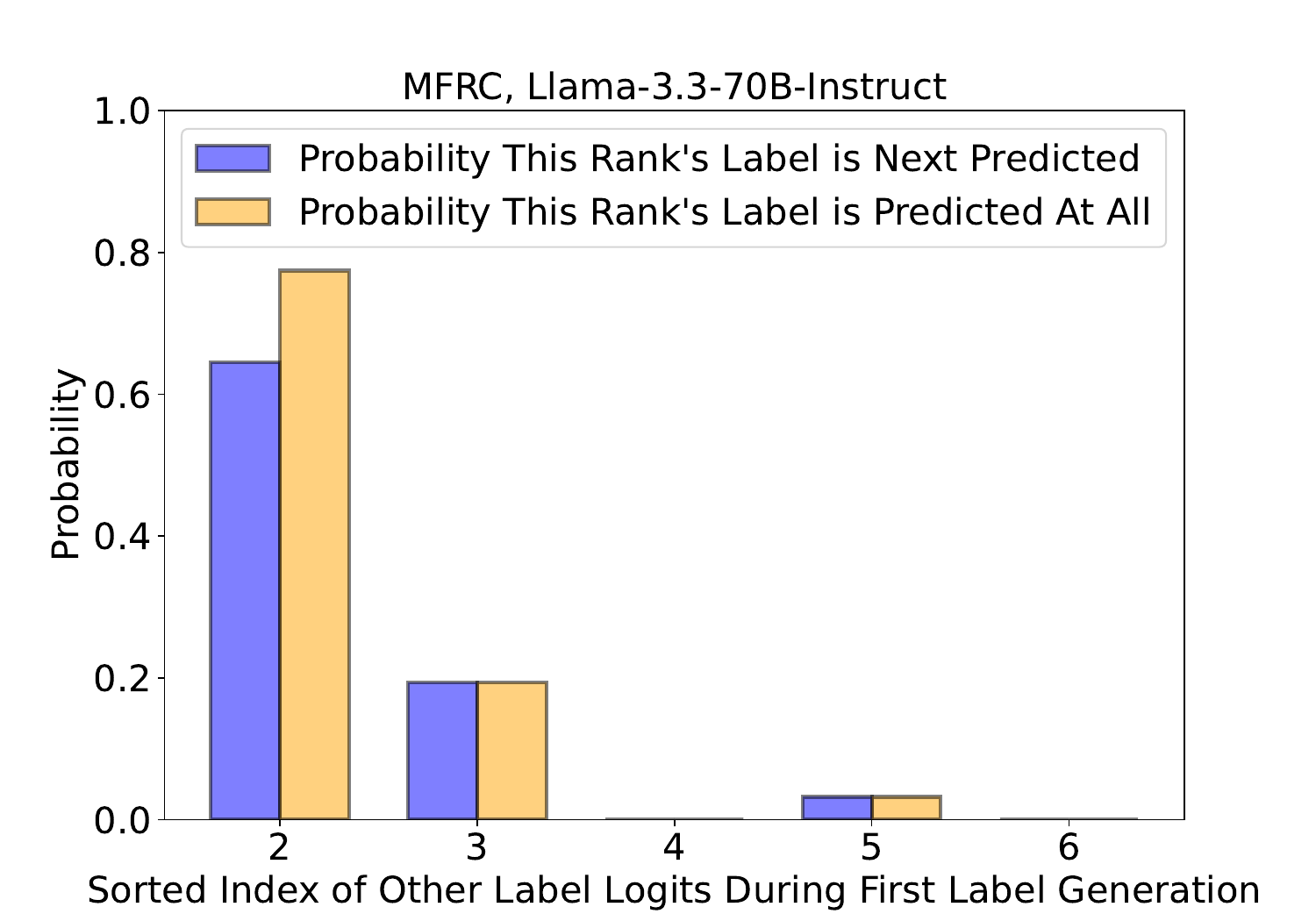}
    
    \caption{Comparing if the label probability distribution created while generating the first label is indicative of what the model will actually predict for multi-label generations on MFRC for Llama-3.1-8B (top) and Llama-3.3-70B (bottom). The first index value is not shown as this corresponds to the actual first label being generated.}
    \label{fig:second_label}
\end{figure}

In Figure~\ref{fig:second_label}, we study in more detail the consistency of the second-highest probability label, excluding the instances where it was not predicted at all, and show the histograms for each generation step. We find that increasing the model size improves the rate at which that label is predicted right after it is ranked second, as Llama3 70B Instruct predicts the label with the second-highest probability as the second label $65$\% of the time compared to approximately $50\%$ of the time with 8B Instruct. This indicates that with scale, the relative ordering of labels improves.

\subsection{Alignment of Llama3 8B} \label{sec:appendix-8b-alignment}

We present results for the alignment of Llama3 8B in addition to the 70B presented in the main text. Results can be seen in Table~\ref{tab:llama_3.1_8B_results}. Our takeaways are virtually identical to 70B, so we refrain from repeating the analysis.

\begin{table*}[!htbp]
\centering
\begin{adjustbox}{width=0.96\textwidth}
    \begin{tabular}{cccccccccccccc}
    \midrule
     & & \multicolumn{6}{c}{Single-Label Datasets}& \multicolumn{6}{c}{Multi-Label Datasets} \\
\cmidrule(lr){3-8} \cmidrule(lr){9-14}
     &   & \multicolumn{3}{c}{Hatexplain} & \multicolumn{3}{c}{MSPPodcast} & \multicolumn{3}{c}{GoEmotions} & \multicolumn{3}{c}{MFRC} \\
\cmidrule(lr){3-5} \cmidrule(lr){6-8} \cmidrule(lr){9-11} \cmidrule(lr){12-14}
    & &  NLL $\downarrow$ & L1 $\downarrow$ & F1 $\uparrow$ & NLL $\downarrow$ & L1 $\downarrow$ & F1 $\uparrow$ & NLL $\downarrow$ & L1 $\downarrow$ & F1 $\uparrow$ & NLL $\downarrow$ & L1 $\downarrow$ & F1 $\uparrow$ \\ \midrule \midrule
    \multirow{2}{*}{\rotatebox{90}{\resizebox{0.9cm}{!}{Baseline}}} & Compare-to-None & \textbf{0.97} & 0.97 & 0.42 & \textbf{1.59} & 1.34 & 0.31 & 33.58 & 5.42 & 0.21 & 20.23 & 4.82 & 0.23 \\
     & Hard Predictions & 12.63 & 1.17 & 0.42 & 13.55 & 1.44 & 0.31 & 27.47 & 1.49 & 0.32 & 40.79 & 2.21 & 0.26 \\

    \midrule
    \multirow{3}{*}{\rotatebox{90}{\resizebox{1.5cm}{!}{Test-Time}}} & Unary Breakdown & 0.98 & 1.01 & 0.35 & 1.62 & 1.48 & 0.12 & 4.99 & 3.21 & 0.29 & 5.29 & 3.03 & 0.22 \\
     & Binary Breakdown & 0.99 & 1.01 & 0.23 & 1.61 & 1.48 & 0.17 & 4.84 & 3.18 & 0.23 & 8.33 & 3.83 & 0.23 \\
     & Max-Over-Generations & \na & \na & \na & \na & \na & \na & 3.00 & 1.44 & 0.34 & 2.87 & 1.58 & 0.39 \\

    \midrule
    \multirow{4}{*}{\rotatebox{90}{\resizebox{1.5cm}{!}{Supervised}}} & BERT & 2.69 & \textbf{0.73} & \textbf{0.66} & 4.29 & \textbf{1.27} & \textbf{0.38} & 2.72 & \textbf{0.63} & \textbf{0.64} & 3.00 & 0.43 & 0.82 \\
     & Linear Probing & \skipped & \skipped & \skipped & \skipped & \skipped & \skipped & \textbf{2.57} & 0.70 & 0.57 & \textbf{2.49} & \textbf{0.39} & \textbf{0.83} \\
     & SFT Outputs & \skipped & \skipped & \skipped & \skipped & \skipped & \skipped & 14.76 & 0.80 & 0.58 & 10.45 & 0.57 & 0.69 \\
     & SFT Max-Over-Generations & \na & \na & \na & \na & \na & \na & 4.15 & 0.72 & 0.57 & 4.87 & 0.54 & 0.73 \\

    \midrule
\end{tabular}
\end{adjustbox}
    \caption{Distribution alignment scores for Llama 3 8B on single and multi-label datasets between LLM and human distributions. $\textbf{F1}\uparrow$ is the example-F1 score. \skipped: Not supplied to avoid clutter.
    % Supervised methods are finetuned on Llama3 8B Instruct (except BERT) on a held-out train set.
    }
    \label{tab:llama_3.1_8B_results}
\end{table*}

\subsection{Effect of Finetuning on Distribution Alignment} \label{sec:appendix-ft-alignment}

\begin{figure}[!htb]
    \centering
    \includegraphics[width=1\columnwidth]{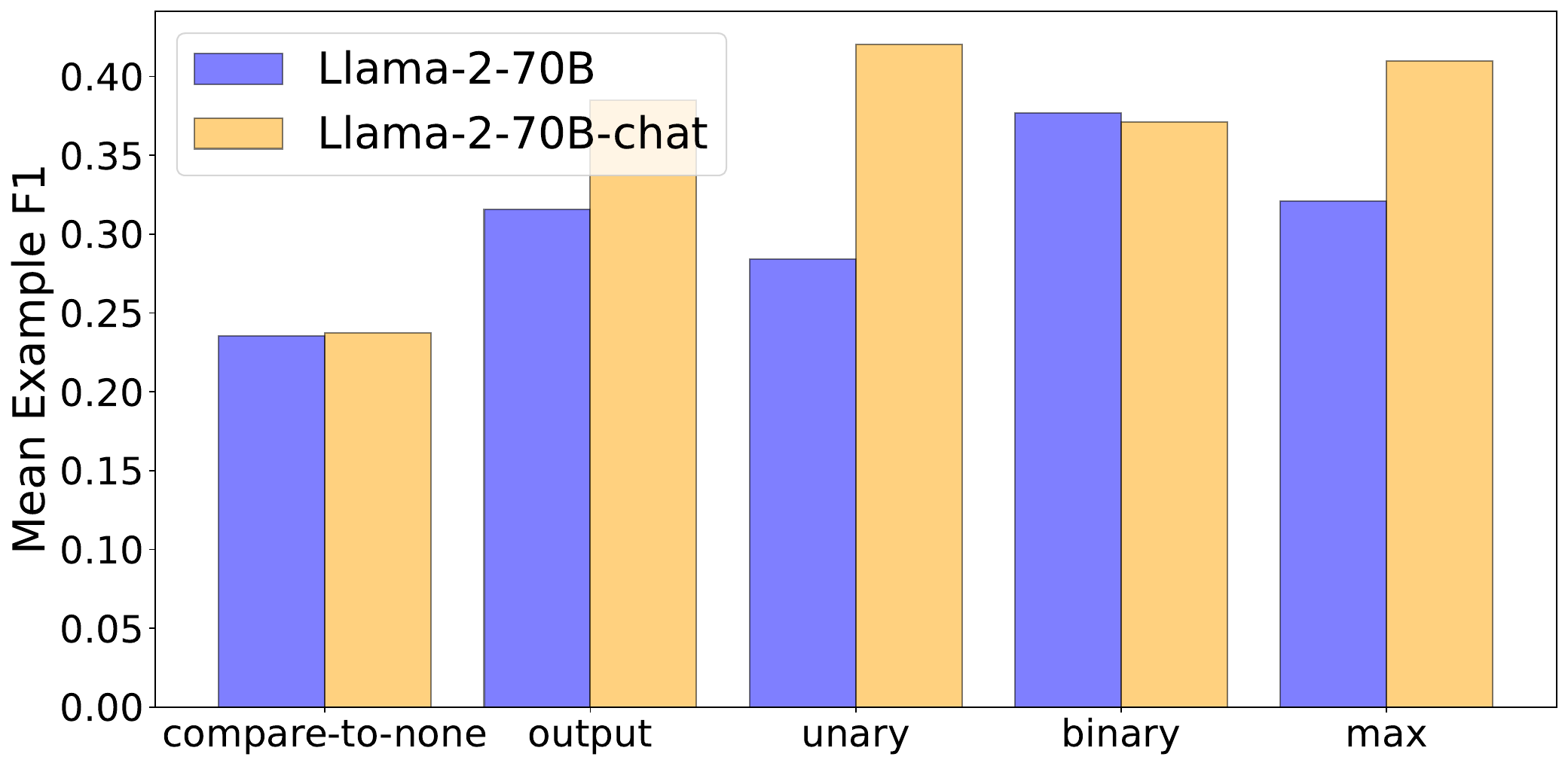}
\includegraphics[width=1\columnwidth]{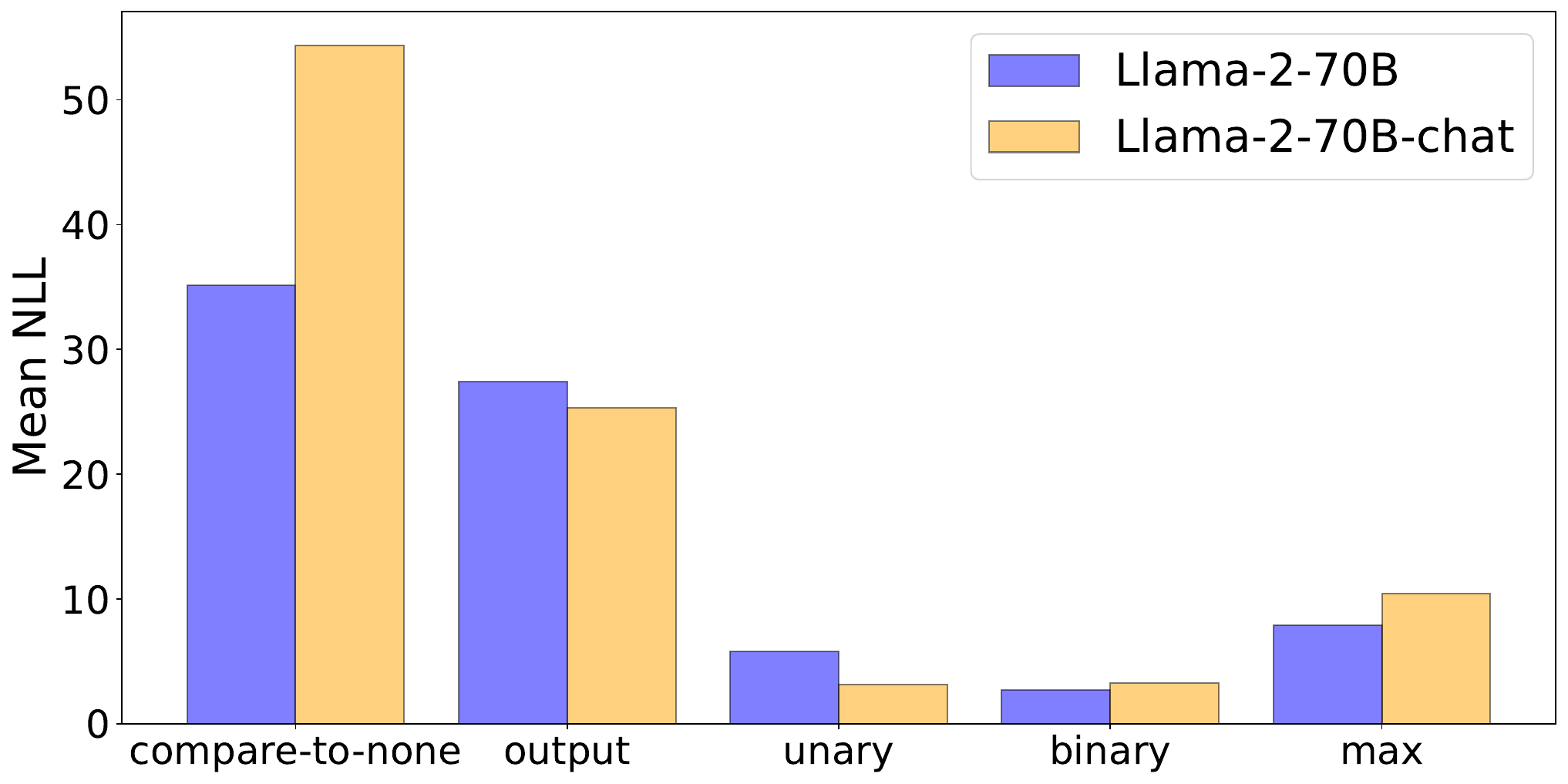}
    
    \caption{Comparing the average example-F1 (top) and Negative Log Likelihood (bottom) between the base Llama-2-70B model and the instruction-finetuned Llama-2-70B-chat model, averaged over MFRC and GoEmotions.}
    \label{fig:chat_comparison}
\end{figure}

Previous research into LLM calibration has found that RLHF \cite{ouyang2022training} can make models more overconfident in their predictions \cite{leng_2025, xie_2024, zhu_2023}. In Figure~\ref{fig:chat_comparison}, we compare the F1 and NLL of Llama-2-70B (base model) and Llama-2-70B-chat (instruction-tuned) for several distribution methods. As expected, the finetuned model generally achieves higher F1 than the base model; however, the NLL for the compare-to-none and max methods (which are the two methods that directly examine the label probabilities) is lower for the base model. This corroborates the aforementioned findings that the model gets more confident when finetuned -- NLL punishes highly confident, wrong answers more than being more confident on correct answers. The similar NLL on unary and binary breakdowns demonstrates that these two methods are relatively robust to different levels of confidence.

\subsection{Attention to Input vs Labels} \label{sec:appendix-attn}

\begin{table*}[!htb]
    \centering
    \begin{adjustbox}{width=0.96\textwidth}

    \begin{tabular}{lcccccccccccc}
        \toprule
        \multirow{2}{*}{Model} & \multicolumn{3}{c}{\textbf{GoEmotions}} & \multicolumn{3}{c}{\textbf{MFRC}} & \multicolumn{3}{c}{\textbf{SemEval}} & \multicolumn{3}{c}{\textbf{Boxes}} \\
        \cmidrule(lr){2-4} \cmidrule(r){5-7} \cmidrule(r){8-10} \cmidrule(r){11-13}
        & \textit{Input} & \textit{Label} & \textit{1st Tokens} & \textit{Input} & \textit{Label} & \textit{1st Tokens} & \textit{Input} & \textit{Label} & \textit{1st Tokens} & \textit{Input} & \textit{Label} & \textit{1st Tokens} \\
        \midrule
        \midrule
        8B Base & 0.242 & 2.04 & 3.62 & 0.132 & 2.01 & 3.29 & 0.162 & 1.76 & 3.00 & 0.095 & 3.11 & 3.06 \\
        8B Instruct & 0.242 & 2.08 & 3.48 & 0.242 & 2.08 & 3.48 & 0.163 & 1.74 & 2.84 & 0.094 & 2.92 & 2.69 \\
        \bottomrule
    \end{tabular}
    \end{adjustbox}
    \caption{Average percentage \% attention to \textit{Input} and \textit{Label} tokens. We also show the average attention to the \textit{1st Tokens} of the labels only, avoiding formatting tokens and the rest of the generated tokens.}
    \label{tab:attn}
\end{table*}

We present the average attention to tokens in the prompt for models, when they generate the second or higher label. We intend to examine how much the models attend to the previous labels generated, establishing empirically the intuition that because of language modeling, the answers of the model deviate from whatever can be gauged from the first generated label token distribution. Table~\ref{tab:attn} shows that, on average, an order of magnitude higher weights are found in the label part of the prompt compared to the input (which also includes other labels because of the demonstrations). Attending to the format of the response is a plausible confounder, so we also check the attention specifically to the first label tokens. This suggests that, indeed, subsequent labels are conditioned on the previous generations. We note that even though average attention is lower on the input, cumulative attention is still greater, with approximately a $80\%$/$20\%$ split in favor of the input, which is usually an order of magnitude or more longer than the labels themselves, again suggesting that a lot of attention weights are accumulated on the generated labels.

\subsection{Results on Qwen} \label{sec:appendix-qwen}

In this section, we replicate our main Llama findings for the Qwen 2.5~\cite{team2024qwen2} family, and in particular for:

\begin{itemize}
    \item Qwen 2.5 7B Base (\texttt{Qwen/Qwen2.5-7B})
    \item Qwen 2.5 7B Instruct (\texttt{Qwen/Qwen2.5-7B-Instruct})
    \item Qwen 2.5 72B Base (\texttt{Qwen/Qwen2.5-72B})
    \item Qwen 2.5 72B Instruct (\texttt{Qwen/Qwen2.5-72B-Instruct})
\end{itemize}

\begin{figure*}[!t]
     \centering
    \includegraphics[width=1\linewidth]{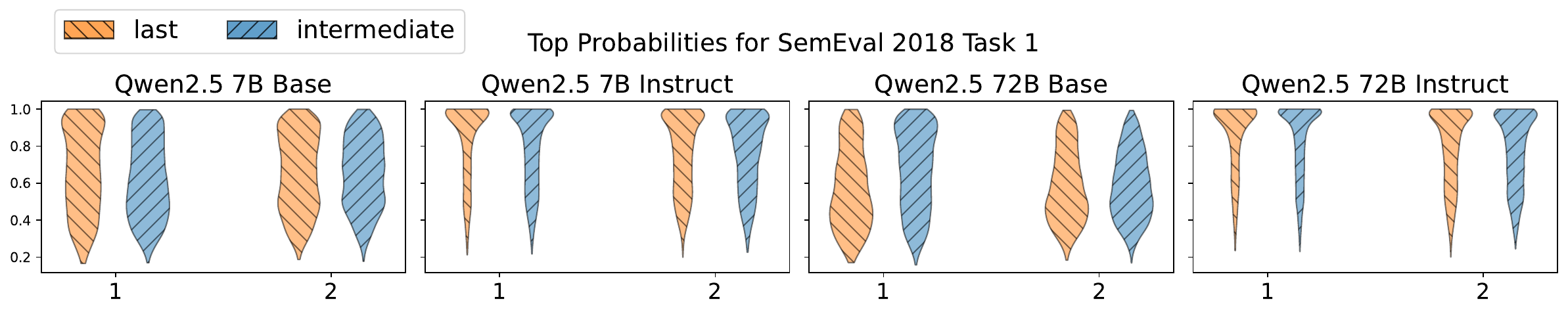}     
    \includegraphics[width=1\linewidth]{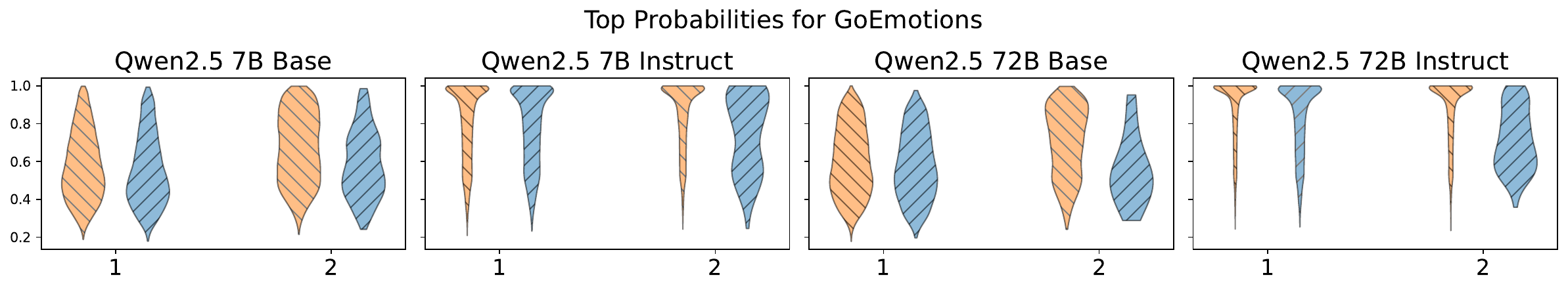}     
    \includegraphics[width=1\linewidth]{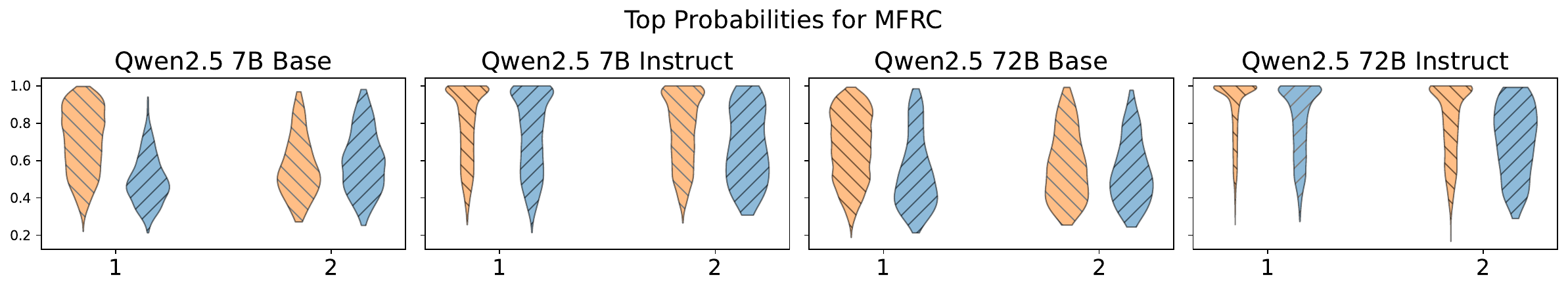}     
    \includegraphics[width=1\linewidth]{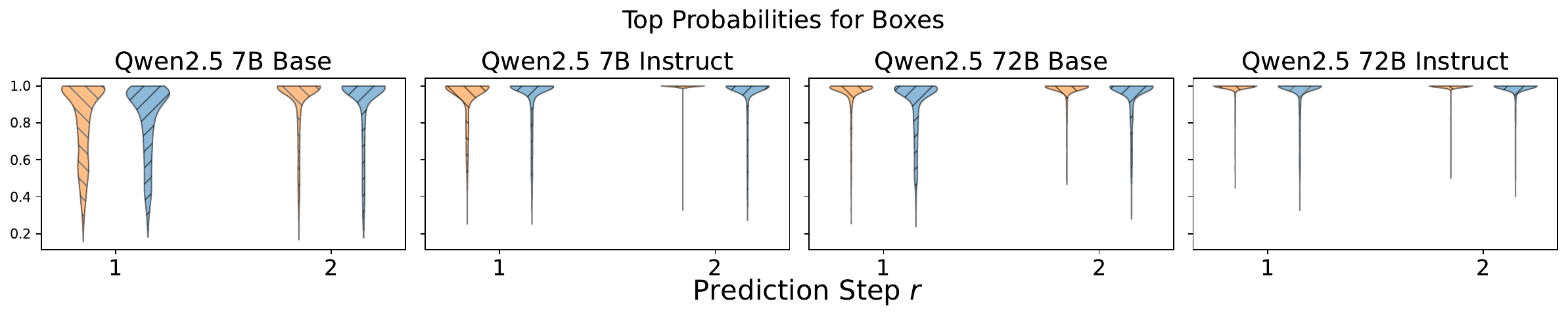}     
    \caption{Top probabilities at each generation step when the \texttt{last} or an \texttt{intermediate} label is generated. Patterns are identical between the two settings, and bigger or finetuned models have clusters closer to 100\%. A single step only is shown when only up to labels were generated for all examples in a specific setting.}
    \label{fig:top-violins-qwen}
\end{figure*}

\begin{figure*}[!t]
     \centering
    \includegraphics[width=1\linewidth]{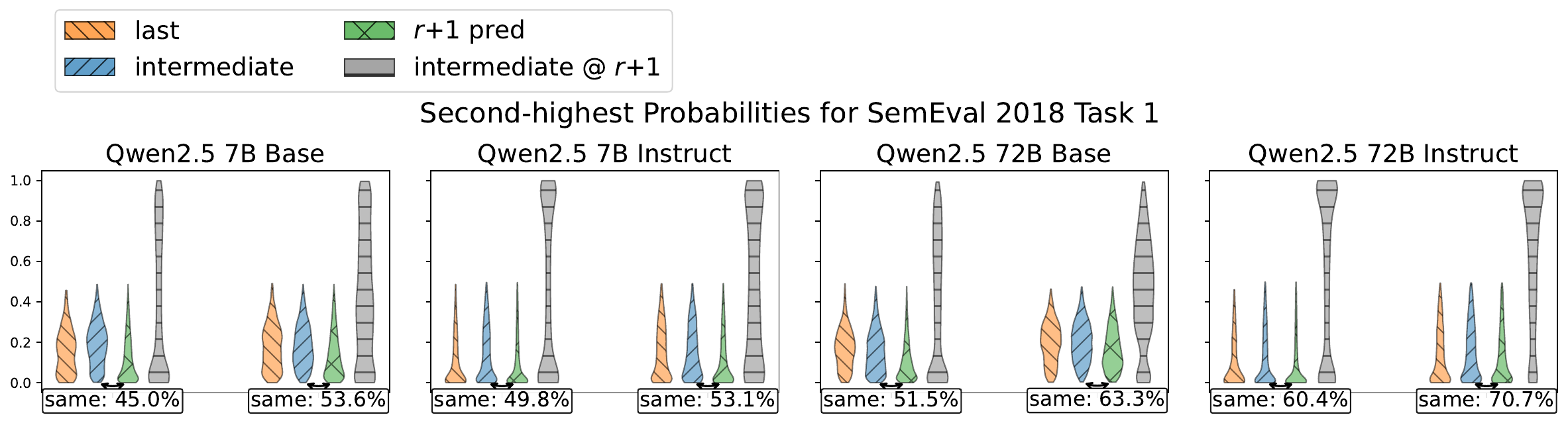}     
    \includegraphics[width=1\linewidth]{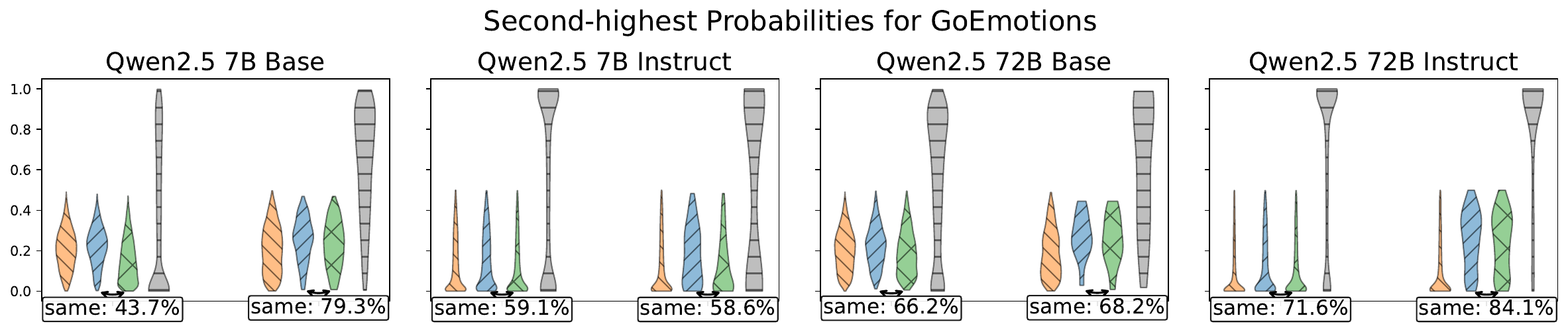}     
    \includegraphics[width=1\linewidth]{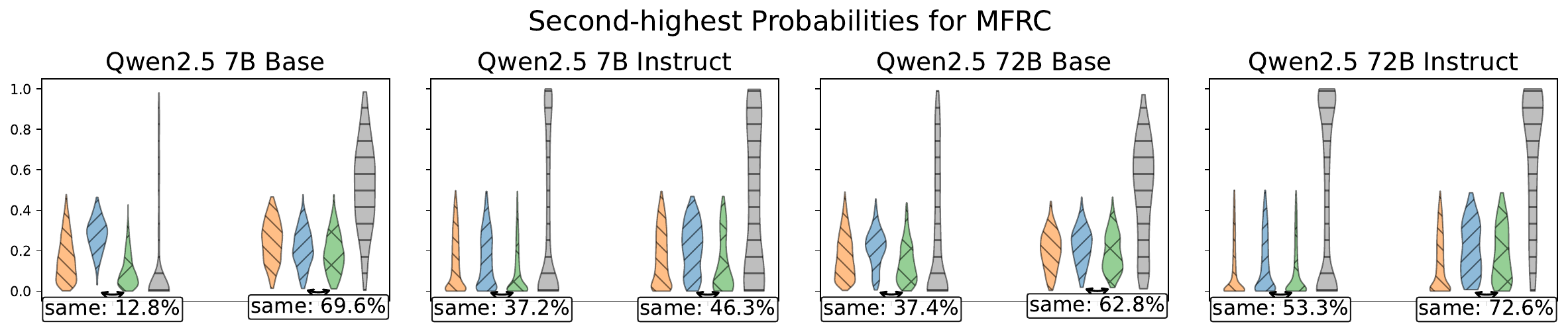}     
    \includegraphics[width=1\linewidth]{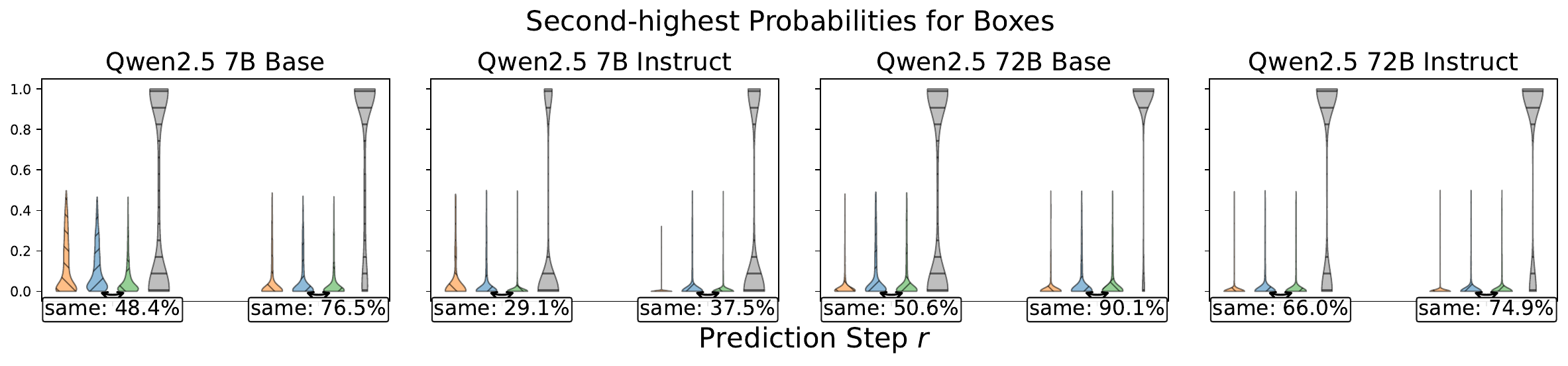}     

    \caption{Second-highest probabilities at each generation step when the \texttt{last} or an \texttt{intermediate} label is generated. We also show the probability at the current step of the label that is actually predicted in the next step ($r$+1 \texttt{pred}), the probability at the next generation step of the second highest probability of the current step (\texttt{intermediate @} $r$+1), and the percentage of cases the second-highest probability label at step $r$ and the prediction at $r$+1 is the \texttt{same}. LLM distributions show poor relative ranking, and little distinction between the \texttt{last} and \texttt{intermediate} settings. A single step only is shown when only up to labels were generated for all examples in a specific setting.}
    \label{fig:second-violins-qwen}
\end{figure*}

\begin{figure*}[!ht]
    \includegraphics[width=1\linewidth]{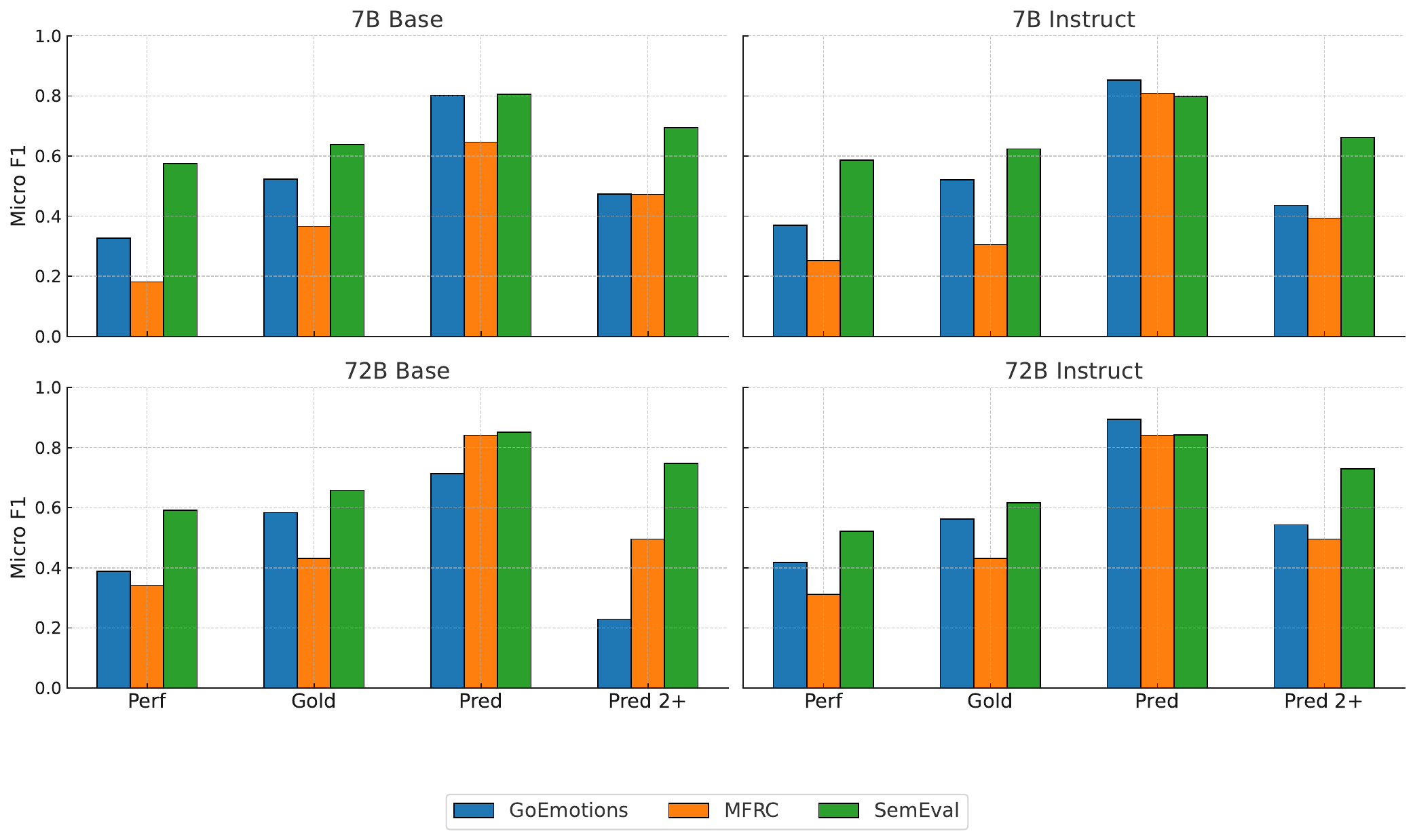}
    \caption{\textbf{Micro F1} $\uparrow$ of linear probes on Qwen 2.5 trained and evaluated on gold labels (\textit{Gold}), trained and evaluated on model predictions (\textit{Pred}), and evaluated on predictions beyond the first generated label (\textit{Pred 2+}). For comparison, we also show the performance of the model (\textit{Perf}). Embeddings are from the last layer for the first generated label.}
    \label{fig:lin_probe_qwen}
\end{figure*}

We present our results for the top two probabilities at each step in Figures~\ref{fig:top-violins-qwen} and \ref{fig:second-violins-qwen}, and our linear probing results in Figure~\ref{fig:lin_probe_qwen}. We see identical with the Llama family, and Qwen can even be said to be more spiky.

\subsection{Medusa: Multiple Decoding Heads} \label{sec:appendix-medusa}

\begin{figure*}[!t]
     \centering
    \includegraphics[width=0.65\linewidth]{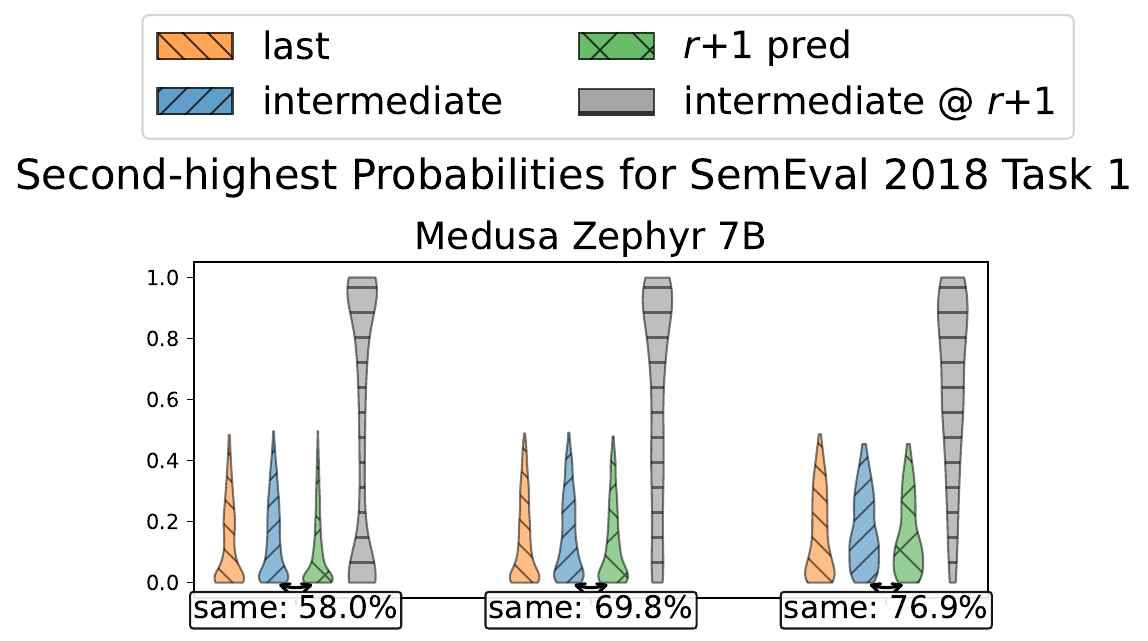}
    \includegraphics[width=0.5\linewidth]{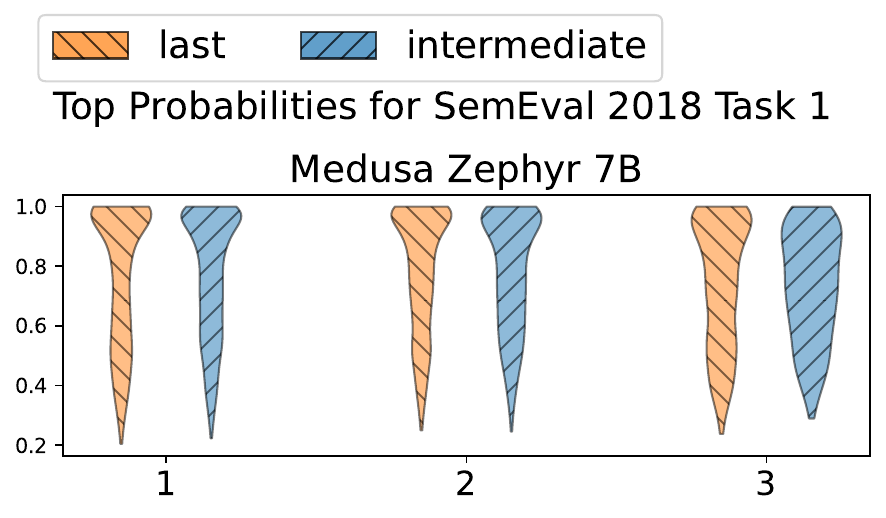}

    \caption{Top and second-highest probabilities at each generation step when the \texttt{last} or an \texttt{intermediate} label is generated. We also show the probability at the current step of the label that is actually predicted in the next step ($r$+1 \texttt{pred}), the probability at the next generation step of the second highest probability of the current step (\texttt{intermediate @} $r$+1), and the percentage of cases the second-highest probability label at step $r$ and the prediction at $r$+1 is the \texttt{same}. Patterns are identical between the two settings, and bigger or finetuned models have clusters closer to 100\%. LLM distributions show poor relative ranking, and little distinction between the \texttt{last} and \texttt{intermediate} settings. A single step only is shown when only up to labels were generated for all examples in a specific setting.}
    \label{fig:violins-medusa}
\end{figure*}

In this section, we present some results from a model with multiple decoding heads, Medusa (specifically \texttt{FasterDecoding/medusa-1.0-zephyr-7b-beta}). Shown in Figures~\ref{fig:violins-medusa}, we see that this model shows behavior similar to Llama 3 and Qwen 2.5.

\subsection{Alphabetical Order} \label{sec:appendix-label-order}

One potential confounding factor in the generation of many labels is their alphabetical order. By default, the labels are presented in an alphabetical order in the instructions and in the demonstrations, as any other mode of presentation would require justification. However, the strong alphabetical priors of the models coupled with the presentation in alphabetical order might be a strong driver of the phenomena we see. Therefore, in this section we present an analysis on how often that happens, as well as randomizing the order of the labels and examining whether the labels follow an alphabetical order or the new order of the instructions. Results are shown in Table~\ref{tab:label-order}, aggregated across Llama3 8B Base and Instruct. We see that proper alphabetical order ossifies the predictions of the model, but the reverse alphabetical order, which is also a regular patter, also does the same yet to a lesser extend. Future research can examine whether randomizing the prompt and aggregating across different orders might help to extract probabilities from the first logits, but this still requires multiple runs, making it more expensive that \textit{Max-over-Generations}.

\begin{table}[]
    \centering
    \begin{tabular}{lcc}
        \textbf{Setting} & \textbf{Alphabetical} (\%) & \textbf{Prompt} (\%) \\
        \midrule
        Alphabetical & 96.4 & - \\
        Random & 35.2 & 40.1 \\
        Reverse & 15.9 & 71.9 \\
    \end{tabular}
    \caption{Percentage of predictions that follow alphabetical order and the order of the labels in the instructions in three settings: \textit{Alphabetical} order of labels, \textit{Random} order of labels (3 different seed) and \textit{Reverse} alphabetical order of labels.}
    \label{tab:label-order}
\end{table}

\end{document}